\documentclass[10 pt, journal, twoside]{IEEEtran}
\IEEEoverridecommandlockouts

\usepackage[T1]{fontenc}

\usepackage[space, compress, sort]{cite}
\usepackage{url}
\usepackage[linkcolor=black, citecolor=black, urlcolor=black, colorlinks=true]{hyperref}

\usepackage{amssymb, amsfonts, amsthm, mathtools, bm}

\usepackage{array} %

\usepackage[ruled,vlined]{algorithm2e}
\usepackage{algpseudocode}
\SetKwRepeat{Do}{do}{while}

\usepackage{graphicx}
\usepackage[font=footnotesize]{caption}

\usepackage{subfigure}
\usepackage{wrapfig}
\usepackage{float}
\usepackage{cuted}  %

\usepackage{tabularx, array, multirow, booktabs, makecell}

\usepackage{enumerate}
\usepackage{verbatim}
\usepackage[normalem]{ulem}   %
\usepackage{xcolor}

\usepackage{pifont}
\definecolor{forestgreen}{RGB}{34,139,34}
\definecolor{firebrick}{RGB}{178,34,34}

\newtheorem{definition}{Definition}
\newtheorem{problem}{Problem}

\newcommand{\delete}[1]{\bgroup\markoverwith{\textcolor{red}{\rule[0.5ex]{2pt}{0.4pt}}}\ULon{#1}}

\usepackage{lipsum}

\newcolumntype{P}[1]{>{\centering\arraybackslash}p{#1}}


\title{Learning Smooth SE(3) Trajectories under Left-Invariant Riemannian Metrics}
\author{Yuwei Wu and Vijay Kumar
\thanks{We gratefully acknowledge the support of The Institute for Learning-Enabled Optimization at Scale (TILOS) funded by NSF Grant CCR-2112665. The authors are with the GRASP Laboratory, University of Pennsylvania, Philadelphia, PA, 19104 USA {\tt\small\{yuweiwu, kumar\}@seas.upenn.edu}.}
 }

\begin{document}
\maketitle
\thispagestyle{empty}

\begin{abstract}
Optimal trajectory generation for rigid-body motions on Lie groups can be formulated as a variational problem that minimizes energy functionals defined by Riemannian metrics. While closed-form solutions exist for special cases such as product metrics and rest-to-rest boundary conditions, solving the general problem with arbitrary boundary states and coupled rotational-translational metrics often requires computationally expensive numerical boundary value solvers. These limitations restrict the use of geometrically consistent trajectory generation in real-time robotic planning and control. This paper presents a learning-based framework for approximating higher-order smooth trajectories on $SE(3)$ under general left-invariant Riemannian metrics. The method parameterizes body-twist trajectories using high-order polynomials and relies on a neural network to learn a subset of the polynomial coefficients and the trajectory duration. The remaining coefficients are analytically determined to enforce the boundary conditions. The training of the network is guided by losses derived from Euler-Lagrange optimality conditions, metric-weighted smoothness objectives, and feasibility constraints. The metric-conditioned framework enables generalization across diverse metric structures and motion conditions. Extensive numerical experiments demonstrate that the proposed approach generates smooth trajectories that closely approximate solutions from numerical optimization while achieving millisecond-level inference times. We demonstrate two practical applications of the proposed framework: real-time generation of diverse motion primitives with waypoint traversal, and refinement for quadrotor flight under dynamic conditions. These results suggest that learning-based motions with geometric structure can provide an efficient alternative to conventional optimization-based methods for trajectory generation on $SE(3)$.
\end{abstract}

\begin{figure}[!ht]
\includegraphics[width=0.99\columnwidth]{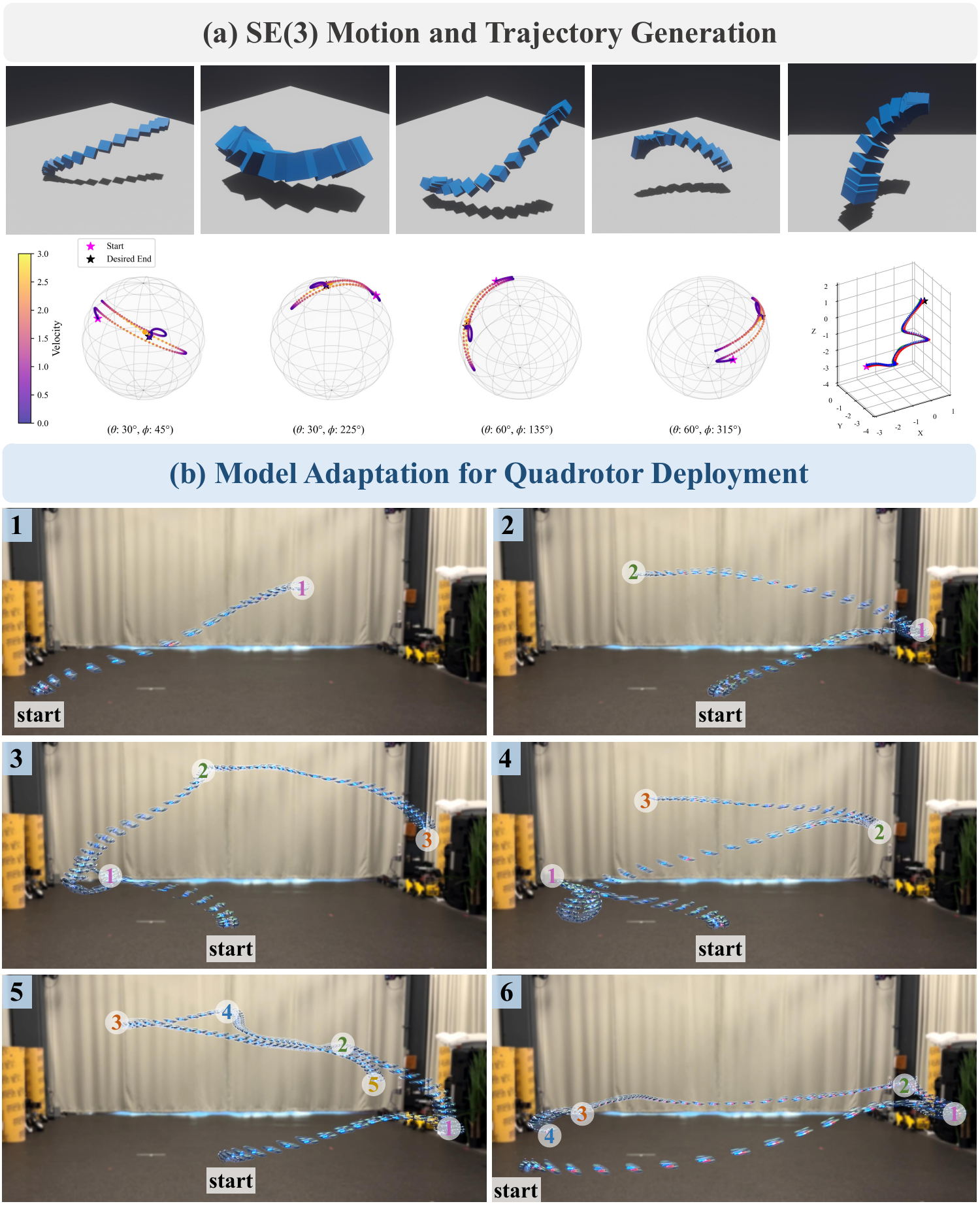}
    \caption{$SE(3)$ trajectories with different boundary conditions and metrics generated by the proposed framework.
    (a) Generation of single motion primitives (first row) and trajectories traversing multiple waypoints with higher-order boundary conditions (second row).
    (b) Fine-tuning adapted for quadrotor applications, evaluated over six different experiments with randomly generated waypoints.}
    \vspace{-0.6cm}
    \label{fig: fig1}
\end{figure}

\section{Introduction}
The problem of generating smooth trajectories that connect initial and terminal states for rigid-body motions on manifolds, especially special orthogonal groups $SO(n)$ and special Euclidean groups $SE(n)$, has been extensively studied over the past decades and has found applications in robotics and related fields~\cite{411534, crouch1995dynamic, 704225,  933127, 1019463, giambo2004optimal}.
Formally, this class of problems can be expressed as nonlinear boundary value problems (BVPs) defined on smooth manifolds, which often require computationally expensive numerical methods due to the underlying geometric and topological complexity of the manifold structure.
Further, coordinate-free formulations and solutions are challenging, as explicitly representing parameterized curves directly on manifolds using a coordinate-free approach is inherently complex~\cite{POPIEL2007111, 10.1093/imamci/12.4.399}.

To address these challenges, several alternative strategies have been developed. 
These include techniques such as singular value decomposition (SVD)-based projection~\cite{1019463}, Bi-Jacobi field methods~\cite{10.4310/CMS.2016.v14.n1.a3}, and generalizations of the De Casteljau algorithm to curved spaces~\cite{DeCasteljau, BoMoVe2018}, each of which leverages different structural properties of the manifold to construct feasible trajectories.
Despite their theoretical appeal, many of these approaches suffer from high computational complexity or restrictive assumptions, which limit their scalability and practical use in real-time robotic applications. 
Existing approaches either rely on restrictive assumptions on the Riemannian metric or require computationally intensive numerical solvers, which limit their use in real-time trajectory generation under general boundary conditions.
As robotic systems become more agile and operate in increasingly complex environments, the growing demand for trajectory generation that is both geometrically consistent and computationally efficient has become increasingly critical.

We focus on trajectory generation for a three-dimensional (3-D) rigid body operating in free space.
The configuration space is modeled as a smooth manifold $\mathcal{M} \subset SE(3)$, the Lie group of rigid body transformations. 
We formulate this as an optimization problem, where the goal is to find a curve \( \gamma: [0, T] \to \mathcal{M} \) that minimizes an energy-related functional,
\begin{subequations}
\label{eq:se3_opt}
\begin{align}
\min_{\gamma, T \in  (0, T_{\text{max}}]} \quad & \mathcal{J}(\gamma, T), \\
\text{s.t.} \quad 
&\nabla_{\gamma}^k \gamma(t) \big|_{t=0} = \gamma_s^{(k)}, \quad \forall k = 0, \ldots, s-1,  \label{eq:manifold_boundary1}  \\
& \nabla_{\gamma}^k \gamma(t) \big|_{t=T} = \gamma_e^{(k)}, \quad \forall k = 0, \ldots, s-1,  \label{eq:manifold_boundary2}  \\
& \mathcal{G}(\gamma, t) \preceq 0,  \quad \gamma(t) \in \mathcal{M}, \quad t \in (0, T]\label{eq:manifold_constraints}
\end{align}
\end{subequations}
where Eq. \eqref{eq:manifold_boundary1} and \eqref{eq:manifold_boundary2} impose boundary conditions on the trajectory $\gamma(t)$ through its covariant derivatives at the initial and terminal time. 
Here, \( \nabla_{\gamma}^k \gamma(t) \) denotes the \( k \)-th order covariant derivatives interpreted on the manifold. The parameter \( T_{\text{max}} \) defines the maximum allowable trajectory duration, and \eqref{eq:manifold_constraints} encodes additional task-specific or feasibility constraints.
The inherent complexity of this problem makes a general solution intractable, motivating the development of alternative methods under simplified conditions.

(1) \textbf{Euclidean embedding} (\( \gamma \in \mathbb{R}^n \)): 
Trajectory optimization on manifolds is often simplified by projecting the problem into Euclidean space, where conventional optimization methods are directly applicable.
This is typically done using local coordinate charts or embeddings that model the manifold as a subset of \( \mathbb{R}^n \)~\cite{watterson2018trajectory}.  
Differentially flat systems~\cite{FLIESS01061995} represent states and inputs as functions of flat outputs in Euclidean space, which can be parameterized by piecewise polynomials~\cite{5980409}, or motion primitives~\cite{7299672, 8206119}, enabling smooth and computationally efficient trajectory generation.

(2) \textbf{Boundary conditions with zero derivatives} ($\nabla_{\gamma}^k \gamma(t) = 0 , k \geq 1 $):
   When all higher-order derivatives at the initial and terminal states are zero, trajectory generation is simplified for rest-to-rest motions such as aerial takeoff, landing, or pick-and-place tasks. 
   Under these conditions,~\cite{704225} showed that trajectories with boundary velocities and accelerations either zero or aligned with the geodesic direction can be generated by reparameterization on the manifold.
    The work in~\cite{7128399} derived the Hamiltonian function of reduced flatness-based dynamics to generate primitives and provided guarantees for non-rest-to-rest transitions.
    These simplifications enable dynamically feasible trajectories with boundary satisfaction, making them well-suited for precomputed motion libraries.
    
(3) \textbf{Specified duration} ($T$):  
Jointly optimizing both the trajectory and its duration introduces significant nonlinearity and increases the complexity of trajectory optimization problems~\cite{10412114}. 
To mitigate this, many methods fix the duration and convert the problem into an optimal-control form, where the objective is to find the optimal trajectory with a fixed time horizon.
To optimize time allocation, other approaches~\cite{5980409, 9147300, 7839930, tordesillas2021faster} decoupled trajectory parameters from timing using bilevel optimization or fixed the path first and adjusted the time-scaling profile to satisfy dynamic constraints~\cite{5256286}.

While effective in limited cases, these strategies 
do not adequately account for intrinsic coupling in the manifold metric and system dynamics under arbitrary boundary conditions, especially for \textbf{non-static} motions.
Hence, generating time-optimal trajectories on manifolds remains intractable for most robotic systems.

To bridge this gap, we introduce a structured learning-based framework for approximating solutions of higher-order variational trajectory optimization problems on \(SE(3)\) by generating Riemannian polynomials that minimize energy functionals under general left-invariant metrics and higher-order boundary conditions.
Trajectories are parameterized in the twist space by high-order polynomials, while smoothness is enforced through a manifold-aware loss derived from the Euler-Lagrange equations. 
Boundary conditions are incorporated via polynomial coefficient completion by solving the linear system, and the loss further penalizes deviations from the target end pose in \(SE(3)\) as well as violations of higher-order derivatives. 
We evaluate the proposed framework through extensive numerical benchmarks and real quadrotor hardware experiments.
Although exact satisfaction of end states in \(SE(3)\) is not guaranteed, the method remains effective when multiple intermediate states are specified, producing trajectories that remain close to the desired conditions while minimizing energy. 
The resulting trajectories are smooth, dynamically feasible, and suitable for real-time replanning, also serving as high-quality initializations for motion planners.
To summarize, our contributions are as follows:
\begin{itemize}

\item We formulate higher-order Riemannian trajectory generation on $SE(3)$ as a structured analytic-learning decomposition in which boundary conditions are enforced exactly via coefficient completion, and the undetermined parameters and trajectory duration are learned.

\item We introduce a metric-conditioned learning framework that generalizes across different left-invariant Riemannian metric families and arbitrary boundary conditions.

\item We demonstrate real-time trajectory generation with millisecond-level inference, achieving high-fidelity approximation to optimization-based solutions, validated through numerical benchmarks and quadrotor hardware experiments with online replanning.

\end{itemize}

\section{Related Works}

\subsection{Trajectory Planning for Rigid Bodies}

Trajectory generation and waypoint interpolation are fundamental to achieving full-body motions in robotic systems. 
The main challenge lies in the manifold structure of the configuration space, which imposes nonlinear constraints that cannot be properly handled by simple Euclidean projection or linear interpolation methods.
For instance, to handle the geometric constraints of $SE(3)$, various approximation methods have been explored.
In~\cite{5f032ddf-06b1-30a4-8397-578b4b7be0ad}, an equal distance projection method was used to fit data onto curves on the sphere by minimizing the projection error.
A related approach in~\cite{1019463} parameterized the trajectory in an ambient matrix space and then applied singular value decomposition (SVD)-based projection to map rotations back to the $SO(3)$ group. 
The selection of orientation representation for two-point interpolation was discussed and compared in~\cite{4608619}, which showed that quartic splines on unit quaternions have smaller approximation errors.

When extending to trajectory generation, several methods formulate cost-minimization problems that are solved by decomposing the optimization or mapping it into reduced spaces.
The work in \cite{BonalliBylardEtAl2019} formulated trajectory optimization as a sequential convex programming approach on manifolds and transformed the original problem into an equivalent Euclidean space formulation.
Similarly, Wang et al.~\cite{9765821} formulated a general nonlinear optimization framework and employed the Hopf fibration~\cite{10.1007/978-3-030-28619-4_20} to map constraints from flat outputs to the original system dynamics.
Watterson et al.\cite{watterson2018trajectory, Watterson-IJRR-2020} generated polynomials on manifolds by leveraging multiple chart transitions and introduced safe corridors on manifolds to ensure obstacle avoidance. 
However, this approach requires finding an initial path on the manifold and depends on the careful selection of parameterizations.
For direct manifold optimization,~\cite{teng2025riemannian} proposed a framework on matrix Lie groups using discrete mechanics to preserve geometric structure while optimizing state and control trajectories without projection.

\vspace{-0.1cm}
\subsection{Learning Approaches for Autonomous Navigation}

Learning-based methods have recently shown significant promise for trajectory planning in robotics, particularly for tasks such as waypoint traversal and autonomous navigation. 
Deep reinforcement learning has demonstrated the ability to learn near-optimal end-to-end control policies for complex drone racing scenarios, enabling high-speed agile flight under time and safety constraints~\cite{song2023reaching, kaufmann2023champion}.
Similarly, imitation learning has been leveraged to train models that generate dense waypoint sequences for downstream tracking with model predictive control (MPC)~\cite{loquercio2021learning}.

However, these methods frequently rely on simplified control commands, lack global trajectory smoothness, often require separate conventional tracking modules, and continue to face challenges such as limited planning horizons and the sim-to-real gap. 
To address these limitations, recent research has shifted toward structured trajectory learning, which aims to generate long-horizon, smooth, and constraint-satisfying trajectories.
For example,~\cite{8593536, pmlr-v205-ryou23a} formulated waypoint traversal as a time-optimal problem using semi-supervised approaches to ensure constraint satisfaction and improve global trajectory quality. 
Similarly, the work in~\cite{pmlr-v211-tankasala23a} utilized a Transformer-based model to learn the time-allocation problem, thereby eliminating the need for bilevel optimization.
Multi-fidelity RL for optimizing time allocation in replanning, particularly during transitions from movement to rest, is presented in~\cite{ryou2024multifidelityreinforcementlearningtimeoptimal}.
\cite{mao2025sequence} learns to approximate model-based time-optimal path parameterization (TOPP) trained on collision-free geometric paths.
Within optimization-based frameworks,~\cite{10412114} introduces implicit neural layers to directly learn time allocation via polynomial trajectory coefficients, and~\cite{han2025dynamically} extends this approach to general nonlinear optimization.

\subsection{Optimization-Based Neural Networks}

Recent advancements have increasingly focused on integrating domain-specific knowledge directly into neural network architectures, enabling them to address structured problems characterized by constraint satisfaction, cost minimization, and physical feasibility. 
For instance, hard constraints can be enforced by utilizing the structure of the network activation functions. 
Amos et al. \cite{10.5555/3305381.3305396, agrawal2019differentiable} proposed a framework that embeds quadratic programming (QP) problems with hard constraints into neural networks by differentiating through the Karush-Kuhn-Tucker (KKT) conditions.
DC3~\cite{donti_dc3_2021} introduced a hybrid approach that learns a subset of decision variables and completes the remaining ones to satisfy equality constraints, followed by a projection step to enforce inequality constraints. 
~\cite{negiar_learning_2023} learned differentiable solvers with meta-learning.
For constrained learning, the work in ~\cite{min2024hardconstrainedneuralnetworksuniversal} further extended the projection methods that inherently satisfy input-dependent affine and convex constraints with universal approximation guarantees.
Rather than explicitly solving constrained optimization problems with networks, physics-driven approaches such as PINNs~\cite{RAISSI2019686, doi:10.1137/21M1397908, SON2023126424}  solve fixed boundary value problems for a specified dynamical system, whereas they generalize poorly when boundary variations extend beyond narrow ranges.

For applications in trajectory optimization, several works have explored embedding structured constraints directly into learning frameworks. 
Implicit differentiable layers have been widely applied in robotics~\cite{jaquier_learning_2022}, as well as in planning and control tasks such as MPC~\cite{amos2018differentiable} and trajectory optimization~\cite{10412114, han2025dynamically}. 
For instance, RAYEN~\cite{tordesillas2023rayenimpositionhardconvex} adapted the projection-based methods and considers a broader class of convex constraints, such as linear, quadratic, second-order cone (SOC), and linear matrix inequality (LMI) constraints. 
It ensures feasibility by projecting infeasible outputs onto the constraint set using a translation from an interior point of the convex region.
These works demonstrate that neural networks designed with embedded optimization principles offer critical advantages for interpretability, reliability, and generalization.

\begin{figure*}[!t]
      \centering
      \includegraphics[width=2.0\columnwidth]{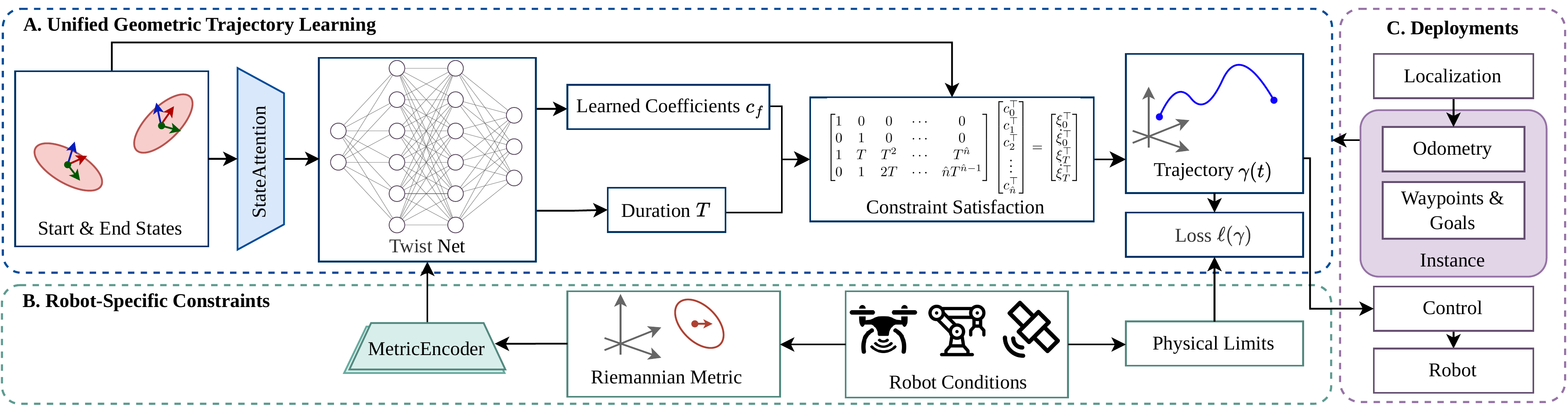}
      \caption{Proposed framework. (A) The learning module predicts the coefficients $c_f$ and the trajectory duration $T$.
      The undetermined coefficients are computed analytically to satisfy boundary constraints. 
      The full trajectory is then evaluated using physically consistent loss functions.
      (B) Different Riemannian metrics and robot dynamics influence the structure of the trajectory and the design of the loss function.
      (C) Given odometry and high-level goals such as waypoints, the network outputs a trajectory that is passed to a low-level controller to produce executable robot actions.}
      \label{fig: network_all}
      \vspace{-0.4cm}
\end{figure*}

\subsection{Learning Under Manifold Constraints}

Neural networks typically rely on a limited set of nonlinear mappings, which can constrain their performance, especially for tasks involving non-Euclidean geometry. 
By directly incorporating manifold structures into neural networks, several approaches have demonstrated how neural networks can be effectively adapted to handle manifold constraints in robotic problems~\cite{duong21hamiltonian, hutchinson2021lietransformer, esteves2018learning}.
For trajectory optimization and path generation, neural networks integrated with manifold representations have been employed to handle high-dimensional and nonlinear dynamics.
Learning from demonstrations on manifolds offers an intuitive and data-efficient way to encode motion patterns while respecting the underlying geometry~\cite{Arvanitidis2021_1000137507, arvanitidis2019fast, 11128556, pmlr-v229-lee23a, 10637485}.
The work in \cite{Arvanitidis2021_1000137507} introduced a method for learning a Riemannian manifold from demonstrations and using graph search to generate an initial trajectory guess. Another data-driven manifold approach in \cite{arvanitidis2019fast} developed a Jacobian-free, fixed-point iterative algorithm to compute the shortest path on the manifold.
To generate continuous trajectories, the work in \cite{10292912} directly learned the nonlinear optimization problem using Lagrangian multipliers and represented the output trajectory as B-spline curves subject to the constrained manifolds.
Recent manifold-based motion primitive approaches either learn a submanifold in the configuration space or a low-dimensional manifold of entire trajectories, with the latter enabling the generation of diverse and adaptable motions~\cite{pmlr-v229-lee23a,10637485}.

However, scaling neural networks to complex manifolds or solving higher-order problems continues to pose challenges, particularly in maintaining stability and interpretability for real-time robotic applications.
We address these problems by proposing a higher-order trajectory learning framework on manifolds that accounts for non-rest states, time scaling, and additional motion dynamics.

\section{Framework Overview}

An overview of the proposed pipeline is shown in Fig.~\ref{fig: network_all}.
The objective is to generate a continuous, differentiable trajectory on the \( SE(3) \) manifold that jointly models rotational and translational motion.
Given the initial and terminal boundary states, the framework parameterizes the body twist trajectory as a polynomial function and predicts the free coefficients together with the total traversal time.
The complete trajectory is then reconstructed by integrating both the learned and analytically derived components, ensuring satisfaction of boundary and continuity constraints.
Training is guided by a sum loss that enforces dynamic consistency, physical feasibility, and smoothness, resulting in improved sample efficiency and stable optimization.
Although trajectories are discretized in time for integration, the underlying dynamics evolve in continuous time, allowing for smooth interpolation and physically consistent motion reconstruction.
Analytically tractable, structure-preserving components are derived using Riemannian geometry, while learning is applied only to undetermined elements, such as trajectory coefficients and duration.

\section{Prerequisites}

\subsection{Lie Group and Lie Algebra}

The configuration of a rigid body in 3-D space, specified by its orientation and position, is represented by an element of the Lie group $SE(3)$, given by
\begin{equation}
SE(3) = \left\{ 
\begin{bmatrix}
R & d \\[2pt]
0 & 1
\end{bmatrix} 
\;\middle|\; R \in SO(3),\; d \in \mathbb{R}^3
\right\},
\end{equation}
where $R \in SO(3) = \{ R \in \mathbb{R}^{3\times 3} \mid R^\top R = I,\; \det(R)=1\}$ is a rotation matrix and $d \in \mathbb{R}^3$ is the position vector.
The associated Lie algebra $\mathfrak{se}(3)$ is
\begin{equation}
\mathfrak{se}(3) = 
\left\{ 
\begin{bmatrix}
\hat{\omega} & v \\[2pt]
0 & 0
\end{bmatrix}
\;\middle|\; \hat{\omega} \in \mathfrak{so}(3),\; v \in \mathbb{R}^3
\right\},
\end{equation}
where $\mathfrak{so}(3) = \{ \hat{\omega} \in \mathbb{R}^{3\times 3} \mid \hat{\omega}^\top = -\hat{\omega}\}$, and the hat operator $\, \hat{\cdot} : \mathbb{R}^3 \to \mathfrak{so}(3)$ maps a vector to its skew-symmetric matrix.
An element of $\mathfrak{se}(3)$ can equivalently be represented by a twist $\{\omega, v\}$, with $\omega \in \mathbb{R}^3$ the angular velocity and $v \in \mathbb{R}^3$ the linear velocity of a chosen reference point in the body frame.
For a smooth trajectory \(X(t): [0, T] \to SE(3)\), the corresponding Lie algebra element is obtained via the tangent map: 
\begin{equation}
    S(t) = X^{-1}(t) \dot{X}(t) = 
\begin{bmatrix}
R(t)^\top \dot{R}(t) & R(t)^\top \dot{d}(t) \\
0 & 0
\end{bmatrix}.
\end{equation}
We adopt the exponential map as a local parameterization from the \( \mathfrak{se}(3) \) to \( SE(3) \), enabling geometrically consistent integration of rigid body motion. When \( S(t) \in \mathfrak{se}(3) \) is constant, the group trajectory is given by $X(t) = \exp\big( t S \big)$.
For a time-varying \( S(t) \), we approximate the trajectory by assuming piecewise-constant twists over small intervals \( \Delta t = T/N\), 
\begin{equation}
X(T) \approx X(0) \cdot \prod_{k=0}^{N-1} \exp\left( \Delta t \cdot S(k \Delta t) \right),
\label{eq: expmap2}
\end{equation}
with the product ordered chronologically. Although \( SE(3) \) is non-commutative, the exponential map and time-ordered integration ensure well-defined trajectory construction.

\begin{definition}
[\textbf{Riemannian Metric on $SE(3)$}]
A Riemannian metric on $SE(3)$ assigns to each $g \in SE(3)$ an inner product 
$\langle \cdot , \cdot \rangle_g$ on $T_g SE(3)$ that depends smoothly on $g$.  

A \textit{left-invariant} metric is determined by its value at the identity $e \in SE(3)$.  
For $S_1, S_2 \in \mathfrak{se}(3)$ with twist coordinates $s_1, s_2 \in \mathbb{R}^6$,  
\begin{equation}
\langle S_1, S_2 \rangle = s_1^\top W s_2, \qquad W \in \mathbb{S}_{++}^6.
\end{equation}
It extends to all $g \in SE(3)$ via left translation:
\begin{equation}
\langle V_1, V_2 \rangle_g 
= \big\langle d(L_{g^{-1}})_g V_1,\, d(L_{g^{-1}})_g V_2 \big\rangle,
\end{equation}
where $V_1, V_2 \in T_g SE(3)$.
\end{definition}

We consider general constant left-invariant metrics defined by 
$W \in \mathbb{S}_{++}^6$, including product and coupled kinetic-energy metrics.
Such metrics arise naturally in geometric mechanics and are widely used in robotics due to their compatibility with rigid-body dynamics.

\begin{definition}[\textbf{Riemannian Connection}~\cite{do1992riemannian}]
Let \( \mathcal{M} \) be a Riemannian manifold equipped with a metric \( g \). The \emph{Riemannian connection}, also known as the \emph{Levi-Civita connection}, is the unique affine connection \( \nabla \) that satisfies:
\begin{itemize}
    \item \textbf{Metric compatibility:} for any vector fields \( X, Y, Z \) on \( \mathcal{M} \),
    \[
    X \cdot \langle Y, Z \rangle_g = \langle\nabla_X Y, Z\rangle_g  + \langle Y, \nabla_X Z\rangle_g ,
    \]
    \item \textbf{Symmetry:} $\nabla_X Y - \nabla_Y X = [X, Y]$,
    where \( [X, Y] \) denotes the Lie bracket of \( X \) and \( Y \).
\end{itemize}
These properties uniquely determine $\nabla$ and ensure compatibility with the manifold's geometry.
\end{definition}
The Riemannian curvature tensor associated with \( \nabla \) measures the noncommutativity of covariant derivatives. For any vector fields \( X, Y, Z \) on \( \mathcal{M} \), it is defined as:
\begin{equation}
\mathcal{R}(X, Y) Z = \nabla_X \nabla_Y Z - \nabla_Y \nabla_X Z - \nabla_{[X, Y]} Z.
\end{equation}
We denote the velocity, acceleration, and jerk along a smooth curve \( X(t) \) as $ V(t) = \dot{X}(t), A(t) = \nabla_V V, 
J(t) = \nabla_V \nabla_V V. $

\subsection{Variational Problems on Manifolds}

Let \( \mathcal{M} \) be a differentiable manifold equipped with a Riemannian metric \( g \). We consider the problem of finding a smooth curve \( X(t): [0, T] \rightarrow \mathcal{M}  \) that minimizes the squared norm of the \( s \)-th order covariant derivative, represented by the following energy functional:
\begin{equation}
\mathcal{J}_{s}(X)
=
\int_{0}^{T}
\left\langle
\nabla_{V}^{s-1}V,
\nabla_{V}^{s-1}V
\right\rangle_{g}
\,dt.
\label{eq: cost_function}
\end{equation}
For a left-invariant Riemannian metric on \(SE(3)\), the inner product
\(\langle\cdot,\cdot\rangle_g\) is induced by the metric matrix \(W\).
The associated Euler-Lagrange equation gives a necessary condition for \( X(t) \) to be a local minimizer~\cite{10.1093/imamci/12.4.399}:
\begin{equation}
\nabla_{V}^{2s-1} V + \sum_{j=2}^{s} (-1)^{j} \mathcal{R}\left( \nabla_{V}^{2s-j-1} V, \nabla_{V}^{j-2} V \right) V = 0.
\label{eq: euler}
\end{equation}
The admissible variations preserve derivatives up to order $s-1$ at the boundary.
We specialize this variational formulation to rigid-body trajectories on
\(SE(3)\).
\begin{definition}[\textbf{Higher-Order Riemannian Variational Curve}]
\label{def:riemannian_variational_curve}
Let $SE(3)$ be equipped with a Riemannian metric \(g\), and let
\begin{equation}
\gamma:[0,T]\rightarrow SE(3), \qquad
\gamma(t)
=
\begin{bmatrix}
R(t) & d(t)\\
0 & 1
\end{bmatrix},
\end{equation}
be a sufficiently smooth curve, where $R(t)\in SO(3)$ and
$d(t)\in\mathbb{R}^3$. Denote its velocity field by
$V(t)=\dot{\gamma}(t)$.

For an integer \(s\geq 1\), a higher-order Riemannian variational curve
is a stationary curve of the functional in Eq.~\eqref{eq: cost_function}, subject to boundary conditions through covariant derivative order \(s-1\).
Equivalently, it satisfies the
Euler-Lagrange equation in Eq.~\eqref{eq: euler} with the boundary conditions as
\begin{subequations}
\label{eq: riepoly}
\begin{align}
\left(
\gamma(0),
V(0),
\nabla_VV(0),
\ldots,
\nabla_V^{s-2}V(0)
\right)
&=
\gamma_{\mathrm{s}}^{[s-1]},
\\
\left(
\gamma(T),
V(T),
\nabla_VV(T),
\ldots,
\nabla_V^{s-2}V(T)
\right)
&=
\gamma_{\mathrm{e}}^{[s-1]}.
\end{align}
\end{subequations}
Here, $\gamma^{[s-1]} := \left(
\gamma, V, \nabla_VV, \ldots, \nabla_V^{s-2}V \right)$ collects  the pose and its covariant derivatives
through order $s-1$.
\end{definition}
For $s=1$, a local minimizer of the functional in Eq.~\eqref{eq: cost_function} corresponds to a geodesic; for $s=2$ and $s=3$, the functional penalizes covariant acceleration and covariant jerk and gives minimum-acceleration and minimum-jerk curves, respectively.
In general, solutions of Eq.~\eqref{eq: euler} are not
polynomials in local coordinates. 
In this work, we approximate these intrinsic variational curves using a finite-dimensional polynomial parameterization of the body twist and instead enforce boundary constraints on the body-twist derivatives.

\subsection{Joint Trajectory Generation}
\begin{figure}[!t]
      \centering
      
 \subfigure[Trajectories with different $T$.]{
\includegraphics[trim={1em 1em 0.5em 4.5em},clip, height=0.37\columnwidth]{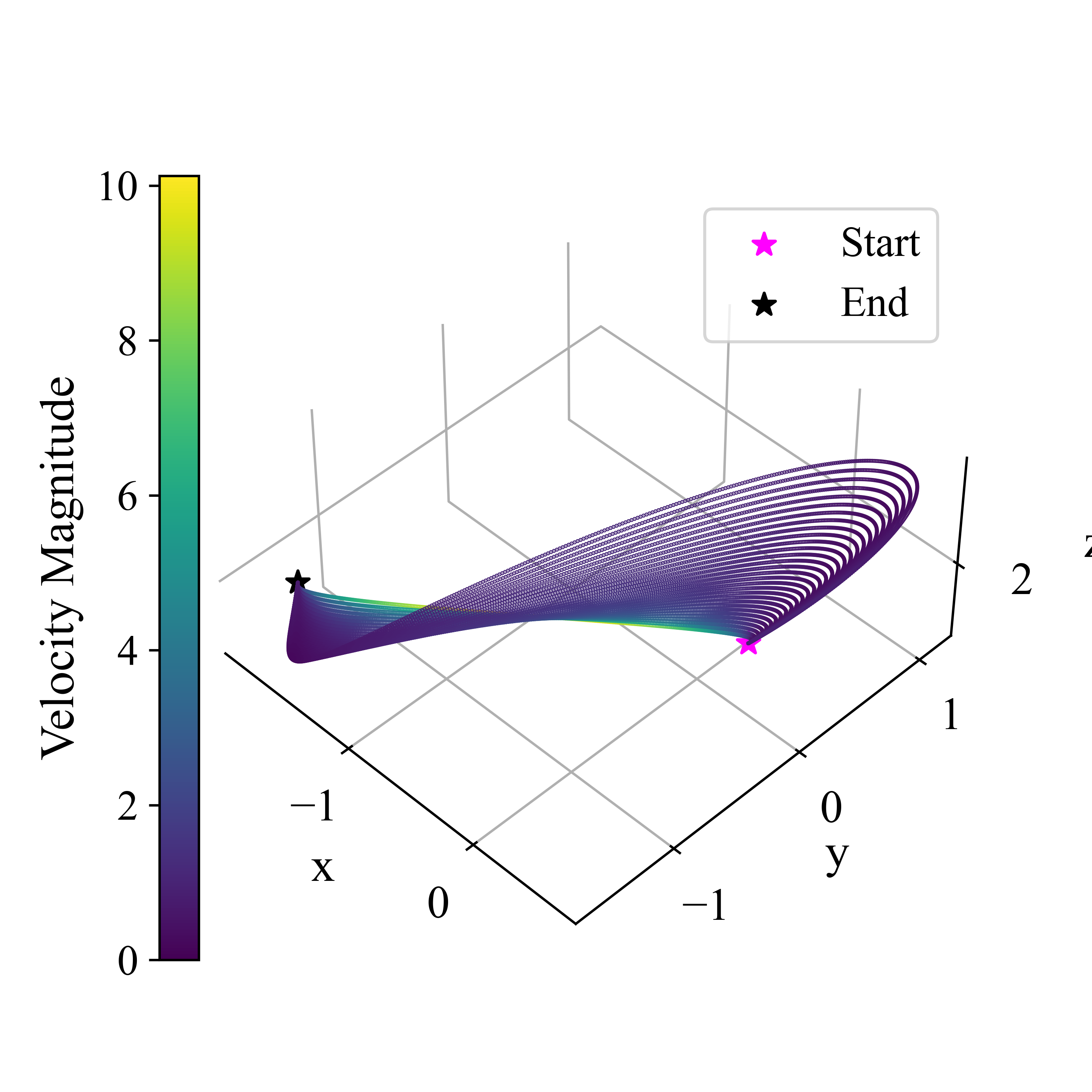}  } 
    \subfigure[Cost function profile.]{
\includegraphics[trim={1em 1em 1em 1em},clip, height=0.37\columnwidth]{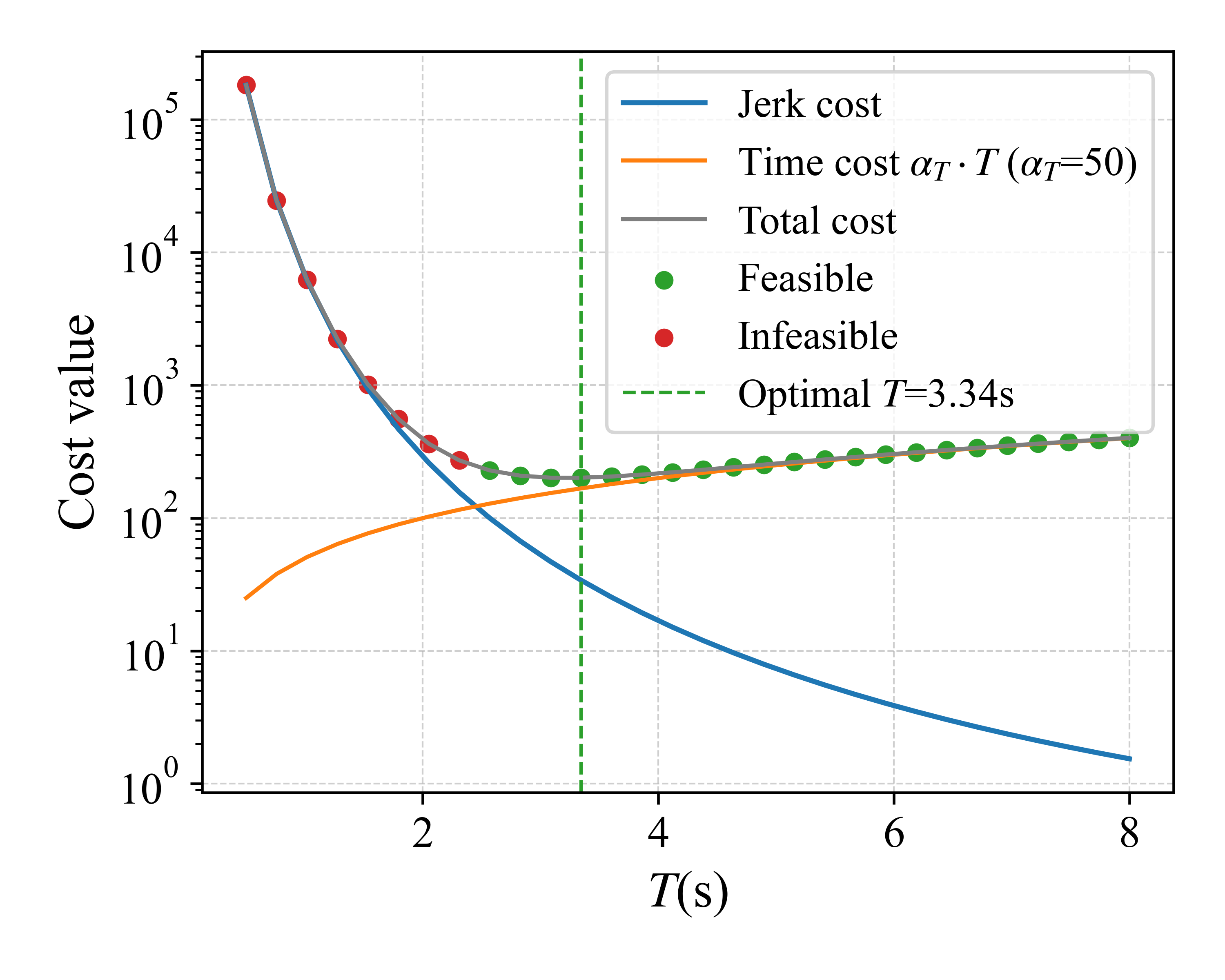}
    }    \vspace{-0.2cm}
      \caption{Time-optimal trajectory generation with trade-off between duration and smoothness in jerk for $\alpha_t $ = 50. Feasibility is defined by the satisfaction of velocity and acceleration limits~\cite{tordesillas2021faster}. 
      }
      \label{fig: time-optimal}
      \vspace{-0.5cm}
\end{figure}

In addition to minimizing the cost in Eq. \eqref{eq: cost_function}, the motion needs to be temporally optimized while respecting actuator and dynamic limits.
\begin{problem}
The joint control and duration trajectory optimization problem can be formulated as
\begin{equation}
\label{prob: timeopt}
\begin{aligned}
\min_{\gamma, T \in  (0, T_{\text{max}}]} \quad
&\mathcal{J}(\gamma, T)=
\mathcal{J}_{m}(\gamma(t), T) + \alpha_t f(T). \\
\text{s.t.}  & \quad  \eqref{eq: riepoly}, 
\end{aligned}
\end{equation}
where $f(\cdot)$ is a time regularization term weighted by $\alpha_t$.
\end{problem}
The time-related function prevents degenerate solutions that the optimizer would drive the duration toward infinity to minimize the first term.
However, it also introduces nonlinearity into the cost functional.
Fig.~\ref{fig: time-optimal} shows an example of one segment trajectory with different durations.
Decoupling the variable to parameterize the time allocation along the path can produce more tractable reformulations for efficient solution with direct optimization or heuristic refinement.

\section{Higher-Order Smoothness Curves on SE(3)}
\label{sec:manifold_formulation}
\subsection{Choices of Left-Invariant Riemannian Metrics}

We begin by considering the Riemannian metrics that arise naturally from the kinetic energy of a rigid body in motion~\cite{doi:10.1177/027836499901800208}.
When the body-fixed frame $\{M\}$ is attached at the center of mass (CoM) and aligned with the principal axes of inertia, the kinetic energy is
\begin{equation}
E_k  = \tfrac12 \langle V, V \rangle 
   = \tfrac12 \bigl( \omega^\top H \,\omega + m\, v^\top v \bigr),
\end{equation}
with
\begin{equation}
W_{H} =
\begin{bmatrix}
H & 0 \\[2pt]
0 & m I_3
\end{bmatrix},  
\end{equation}
where $H = \operatorname{diag}(H_{xx}, H_{yy}, H_{zz})$ containing the principal moments of inertia and $m \in \mathbb{R}_{+}$ is the mass.
Thus, $W_H$ defines a left-invariant Riemannian metric on $SE(3)$. %
If we specialize to the isotropic case by setting $H = \alpha I_3$ and $m = \beta $, then
\begin{equation}
W_0 =
\begin{bmatrix}
\alpha I_3 & 0 \\[2pt]
0 & \beta I_3
\end{bmatrix},    
\end{equation}
yielding an isotropic, scale-dependent metric~\cite{doi:10.1177/027836499401300101}.

Consider a new body frame $\{M\}_C$ obtained from $\{M\}$ by the rigid transform
\begin{equation}
\mathcal{T}_C =
\begin{bmatrix}
Q & c \\ 0 & 1
\end{bmatrix}, \quad Q \in SO(3),\; c \in \mathbb{R}^3.
\end{equation}
Let \( C := [c]_\times \), the kinetic energy becomes
\begin{equation}
\begin{aligned}
E_k &= \tfrac{1}{2} \left(
\omega^\top Q^\top H Q\, \omega + m\, (v + \omega \times c)^\top (v + \omega \times c)
\right) \\
&= \tfrac{1}{2}
\begin{bmatrix}
\omega \\ v
\end{bmatrix}^\top
W_C
\begin{bmatrix}
\omega \\ v
\end{bmatrix},
\end{aligned}
\end{equation}
where
\begin{equation}
W_C =
\begin{bmatrix}
Q^\top H Q - m Q^\top C^2 Q & -\,m Q^\top C Q \\[4pt]
m Q^\top C Q & m I_3
\end{bmatrix}.
\end{equation}
This defines the most general constant left-invariant Riemannian metric on $SE(3)$ induced by rigid-body kinetic energy (known as the spatial inertia matrix).

For geometric modeling or regularization tasks that may penalize translations differently along different axes, we extend to a more general metric
\begin{equation}
W_{B} \;=\;
\begin{bmatrix}
H & 0 \\[2pt]
0 & B
\end{bmatrix},
\qquad B \in \mathbb{S}_{++}^3, \ B\neq m I_3.
\end{equation}
This metric remains a valid constant left-invariant Riemannian metric on $SE(3)$ that anisotropically weights translations.
Similarly, for general $(Q,c)$ transform with anisotropic $B$, the corresponding metric is
\begin{equation}
W_{CB} =
\begin{bmatrix}
Q^\top (H + C^\top B C)\, Q & Q^\top C^\top B Q \\[4pt]
Q^\top B C Q & Q^\top B Q
\end{bmatrix}
\label{eq:W_QcHB}
\end{equation}
We will primarily focus on these metrics that are physically meaningful to general robot motions. 

\begin{figure*}[!t]
      \centering
    \includegraphics[width=2\columnwidth]{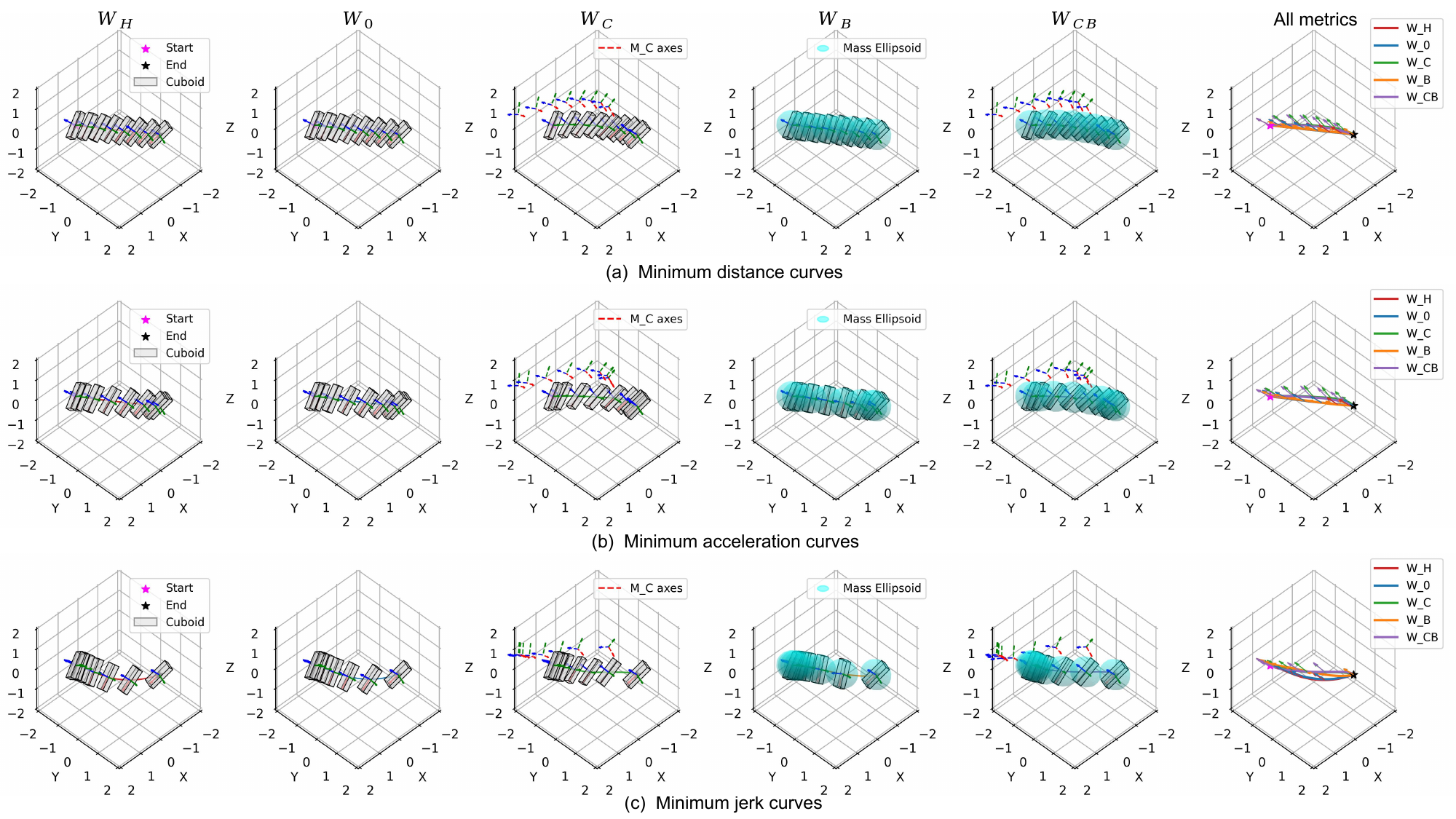}
      \caption{Trajectories optimized using numerical solvers under five distinct Riemannian metrics with different functionals: (a) minimum distance, (b) minimum acceleration, and (c) minimum jerk.}
      \label{fig: variation}
      \vspace{-0.5cm}
\end{figure*}
\subsection{Levi-Civita Connection under Different Metrics}
For vector fields $X=\{\omega_x,v_x\}, Y=\{\omega_y,v_y\}, Z=\{\omega_z,v_z\}$, the Levi-Civita connection with respect to a left-invariant metric $W=\begin{bmatrix}G & D^\top \\ D & B\end{bmatrix} \in \mathbb{S}_{++}^6 $ satisfies
\begin{equation}
\begin{aligned}
\langle Z, \nabla_{X} Y \rangle 
&= \langle Z,\, X(Y^{i}\hat{L}_{i}) \rangle \\
&+ \tfrac{1}{2}\Big[
\langle \{\,\omega_{z} \times \omega_{y},\;\; \omega_{z} \times v_{y} + v_{z} \times \omega_{y}\}, \{\omega_{x}, v_{x}\}\rangle \\
&+ \langle \{\,\omega_{z} \times \omega_{x},\;\; \omega_{z} \times v_{x} + v_{z} \times \omega_{x}\}, \{\omega_{y}, v_{y}\}\rangle \\
&+ \langle \{\,\omega_{x} \times \omega_{y},\;\; \omega_{x} \times v_{y} + v_{x} \times \omega_{y}\}, \{\omega_{z}, v_{z}\}\rangle
\Big] \\[2pt]
&= \langle Z,\,X(Y^i\hat L_i)\rangle \\
&+ \tfrac12 \langle Z,\{\,\omega_x \times \omega_y,\;\; \omega_x \times v_y + v_x \times \omega_y \}\rangle \\
& + \tfrac12 (
\omega_z^\top \Phi_\omega + v_z^\top \Psi_v ),
\end{aligned}
\end{equation}
where
$$
\begin{aligned}
\Phi_{\omega}(X, Y) &= \omega_y \times (G\omega_x+D^{\top} v_x) + v_y \times (D\omega_x+B v_x) \\
&\quad + \omega_x \times (G\omega_y+D^{\top} v_y) + v_x \times (D\omega_y+B v_y), \\[4pt]
\Psi_{v}(X, Y) &= \omega_y \times (D\omega_x+B v_x) + \omega_x \times (D\omega_y+B v_y),
\end{aligned}
$$
Thus, 
\begin{equation}
\begin{aligned}
\nabla_X^{W} Y
&= \begin{bmatrix}
\dot{\omega} +\tfrac{1}{2}  \omega_x \times \omega_y  \\
\dot{v} + \tfrac{1}{2}(\omega_{x} \times v_{y} + v_{x} \times \omega_{y})
\end{bmatrix}\\
& \quad + \tfrac{1}{2}  W^{-1}
\begin{bmatrix}
\Phi_{\omega}(X, Y) \\ \Psi_{v} (X,Y)
\end{bmatrix}
\label{eq: nabla_X}
\end{aligned}
\end{equation}
Since anisotropic cases can be derived as extensions of the isotropic and general kinetic energy metrics, we focus below on detailing three representative metrics.

\subsubsection{Isotropic Metric}
The inner product reduces to
$\langle Z, \{a,b\} \rangle := \alpha a^{\top}\omega_z + \beta b^{\top} v_z,$
with $\Phi_{\omega} = 0, \ \Psi_{v} = \beta ( \omega_y \times v_x + \omega_x \times v_y)$, then
\begin{equation}
\quad \nabla_X^{W_0} Y = \left\{ \dot{\omega}_y  + \tfrac{1}{2}\,\omega_x \times \omega_y,
\dot{v}_y + \omega_x \times v_y \right\}.
\end{equation}

\subsubsection{Centroidal Principal-Axes Metric}
We have
\begin{equation}
\begin{aligned}
\langle Z, \{a,b\} \rangle :&= a^{\top} H \omega_z + m b^{\top} v_z.\\
\Phi_{\omega} &= \omega_y \times (H\omega_x) + \omega_x \times (H\omega_y), \\
\Psi_{v} &= m ( \omega_y \times v_x + \omega_x \times v_y).
\end{aligned}
\end{equation}
Therefore,
\begin{equation}
\begin{aligned}
\nabla_{X}^{W_k} Y = 
\left\{
    \dot{\omega}_y 
    + \tfrac{1}{2}\big[ \omega_{x} \times \omega_{y}
    + H^{-1} (\omega_{x} \times (H \omega_{y})) \right . \\
    \left. + H^{-1} (\omega_{y} \times (H \omega_{x})) \big], \;
    \dot{v}_y + \omega_{x} \times v_{y}
\right\}
\end{aligned}
\end{equation}

\subsubsection{General Kinetic-Energy Metric}

We denote $C_Q := Q^\top C Q = [Q^\top c]_\times$, $K := Q^\top H Q$, $J := K - m\,C_Q^{2}$.
The displaced metric takes the compact form as
\begin{equation}
W_C =
\begin{bmatrix}
J & -\,m C_Q\\[4pt]
mC_Q & m\,I_3
\end{bmatrix},
\end{equation}
and
\begin{equation}
\langle Z, \{a,b\} \rangle := \begin{bmatrix}
z_\omega \\ z_v
\end{bmatrix}^T W_C  
\begin{bmatrix}
a \\ b
\end{bmatrix}.
\end{equation}
Here,
\begin{equation}
\begin{aligned}
\Phi_{\omega} &= \omega_y \times (J\omega_x) + \omega_x \times (J\omega_y) \\
&- m \omega_y \times (C_Q v_x) + m v_y \times (C_Q\omega_x) \\
&- m \omega_x \times (C_Q v_y) + m v_x \times (C_Q \omega_y)
\\[2pt]
\Psi_{v} &= m \omega_y \times (C_Q \omega_x) + m \omega_x \times (C_Q \omega_y) \\
&+ m ( \omega_y \times v_x + \omega_x \times v_y)
\end{aligned}
\end{equation}
Substituting into Eq. (\ref{eq: nabla_X}) gives the final formulation.
If $c=0\Rightarrow C_Q=0$, then $J=K$ to recover the centroidal expression; if further $H=\alpha I_3$, it obtains the isotropic expression.

\subsubsection{Levi-Civita Connection under a Change of Metric}
Let ${}^W\nabla$ be the Levi-Civita connection of a general left-invariant Riemannian metric $W$ on $SE(3)$, and let ${}^{W_0}\nabla$ denote that of a nominal reference (e.g. isotropic) metric $W_0$.
A change in the metric from
\(W_0\) to \(W\) generally induces a corresponding change from
\({}^{W_0}\nabla_X Y\) to \({}^{W}\nabla_X Y\)
\cite{do1992riemannian}, which affects the higher-order covariant derivatives and the resulting optimal trajectory. 
In this work, we utilize this metric dependence by conditioning the trajectory neural network on $W$. 
This allows the model to learn metric-dependent trajectory variations and adapt its parameterization across different metric structures.

\subsection{Necessary Conditions under Different Metrics}
We summarize results from~\cite{704225} and restate the necessary conditions derived from Eq.~\eqref{eq: euler} under different metrics.

\subsubsection{Minimum Distance Curves}
The Euler-Lagrange equation reduces as the covariant acceleration vanishes, i.e., $\nabla_V V = 0$.
For an isotropic metric, the formulation is
\begin{equation}
A = \nabla_V V =
\begin{bmatrix}
\dot\omega \\[2pt]
\dot v+\omega\times v
\end{bmatrix}
 =
\begin{bmatrix}
\dot\omega \\[2pt]
R^T d^{(2)}
\end{bmatrix},
\end{equation}
where $v=R^T \dot{d}$. Similarly, for the centroidal principal-axes metric, 
\begin{equation}
A = 
\begin{bmatrix}
\dot\omega + H^{-1}\big(\omega\times(H\omega) \\
\dot v+\omega\times v,
\end{bmatrix},
\end{equation}
and for the general kinetic-energy metric, we have 
\begin{equation}
A= \begin{bmatrix}
\dot{\omega}  \\
\dot{v}
\end{bmatrix}
+ \tfrac{1}{2}  W_C^{-1}
\begin{bmatrix}
\Phi_{\omega}(V, V) \\ \Psi_{v} (V, V)
\end{bmatrix},
\end{equation}
where
\begin{equation}
\begin{aligned}
\quad \Phi_{\omega}(V, V) &= 2 \omega \times (J\omega)  \\
&- 2 m \omega \times (C_Q v) + 2 m v \times (C_Q\omega)
\\[2pt]
\Psi_{v} (V, V) &= 2 m \omega \times ( C_Q \omega  +  v ).
\end{aligned}
\end{equation}

\subsubsection{Minimum Acceleration/Jerk Curves}
The isotropic case has the closed-form expression for higher-order covariant derivatives.
The jerk and snap are:
\begin{equation}
\begin{aligned}
J
&= \begin{bmatrix}
\omega^{(2)} + \tfrac{1}{2}\,\omega \times \dot{\omega} \\
R^T d^{(3)}
\end{bmatrix},
\\[2pt]
\nabla_{V}^{3} V
&= \begin{bmatrix}
\omega^{(3)} + \omega \times \ddot\omega + \tfrac{1}{4}\,\omega \times (\omega \times \dot\omega) \\[8pt]
R^T d^{(4)}
\end{bmatrix}.
\end{aligned}
\end{equation}
Since the translation is purely the derivative of $d$ and decoupled from $\omega$, the fifth derivative of the rotational part is
\begin{equation}
\begin{aligned}
(\nabla_{V}^{5} V)_\omega  =\; &\omega^{(5)} 
+ 2\, \omega \times \omega^{(4)} 
+ \tfrac{5}{2} \dot{\omega} \times \omega^{(3)} \\
&+ \tfrac{3}{2} \omega \times (\omega \times \omega^{(3)}) + \tfrac{5}{4} \omega \times (\dot{\omega} \times \omega^{(2)}) \\
&+ \dot{\omega} \times (\omega \times \omega^{(2)}) 
+ \tfrac{1}{4} \omega^{(2)} \times (\omega \times \dot{\omega}) \\
&+ \tfrac{1}{8} \dot{\omega} \times (\omega \times (\omega \times \dot{\omega})) 
+ \tfrac{1}{8} \omega \times (\dot{\omega} \times (\omega \times \dot{\omega})) \\
&+ \tfrac{1}{2} \omega \times (\omega \times (\omega \times \omega^{(2)})) \\
&+ \tfrac{1}{16} \omega \times (\omega \times (\omega \times (\omega \times \dot{\omega}))).
\end{aligned}
\end{equation}
For a minimum acceleration curve, the condition yields
\begin{equation}
\begin{aligned}
 \nabla_{V}^3 V + \mathcal{R}(\nabla_{V} V, V) V 
&= \begin{bmatrix}
\omega^{(3)} + \omega \times \ddot\omega \\[6pt]
R^T d^{(4)}
\end{bmatrix}= \mathbf{0}.
\end{aligned}
\end{equation}
Similarly, for a minimum-jerk curve,
\begin{equation}
 \nabla_{V}^5 V + \mathcal{R}(\nabla_{V}^3 V, V) V - \mathcal{R}(\nabla_{V}^2 V, \nabla_{V} V) V= \mathbf{0}.
\end{equation}
Substitution into the necessary condition,
\begin{equation}
\mathcal{E}(\xi^{[5]}) \;=\;
\begin{bmatrix}
\mathcal{E}_1(\omega^{[5]}) \\[4pt]
d^{(6)}
\end{bmatrix}
= \mathbf{0},
\label{eq:boundary}
\end{equation}
where
\begin{equation}
\begin{aligned}
\label{eq: omega1}
\mathcal{E}_1(\omega^{[5]}) = &\omega^{(5)}  +2 \omega \times \omega^{(4)} +\tfrac{5}{2} \dot{\omega} \times \omega^{(3)}  \\
 &  +\tfrac{5}{4} \omega \times(\omega \times \omega^{(3)}) +\tfrac{3}{2} \omega \times(\dot{\omega} \times \omega^{(2)}) \\
 &+ \dot{\omega} \times (\omega \times \omega^{(2)}) + \tfrac{1}{4} \omega^{(2)} \times (\omega \times \dot{\omega}) \\
&+\tfrac{3}{8} \omega \times(\dot{\omega}  \times  (\omega \times \dot{\omega})) +\tfrac{1}{4} \omega \times(\omega \times(\omega \times \omega^{(2)})) \\
&+ \tfrac{1}{8}\dot{\omega} \times (\omega \times(\omega \times \dot{\omega}))
\end{aligned}
\end{equation}
This decouples into rotational and translational parts.
The translation part $d(t)$ can be uniquely determined by boundary conditions via a fifth-degree polynomial, while the rotational dynamics satisfy Eq.~\eqref{eq: omega1}. 
In general, \eqref{eq: omega1} does not have a closed-form solution for arbitrary boundary conditions when $s>1$, though special solvable cases are detailed in~\cite{704225}.

We show a numerical example in Fig.~\ref{fig: variation}, where trajectories are computed under different Riemannian metrics. 
For product metrics $W_H$ and $W_0$ with decoupled rotations and translations, the optimization converges fast.
The metrics $W_C$ and $W_B$ induce nontrivial Levi-Civita connections, producing noticeably different trajectories. 
In particular, the anisotropic metric $W_B$ weights motion differently along various directions of the rigid body, which also leads to curved trajectories that deviate from the geodesic solutions of product metrics.

\section{Learning Structured SE(3) Trajectories}

\subsection{Problem Formulation}

To achieve a more efficient representation and keep the network output size consistent across different discretization resolutions, we approximate the twist trajectory as a higher-order polynomial of degree $n$:
\begin{equation} 
\begin{aligned} 
\label{eq:poly_twist}
&\xi(t) = \sum_{j=0}^n c_j \, t^j = \mathbf{c} \, \beta(t) \in \mathbb{R}^6, \\ 
&\text{where } \ \beta(t) = [1,\, t,\, t^2,\, \ldots,\, t^n]^\top \in \mathbb{R}^{n+1}, \\
&\ \qquad \qquad \mathbf{c} = \begin{bmatrix} c_0 & c_1 & \cdots & c_n \end{bmatrix} \in \mathbb{R}^{6 \times (n+1)} .
\end{aligned} 
\end{equation}
The $k$-th time derivative is given explicitly by
\begin{equation}
\label{eq:poly_deriv}
\xi^{(k)}(t) = \sum_{j=k}^n \frac{j!}{(j-k)!}\, c_j \, t^{\,j-k},
\quad k = 0,1,\ldots, n.
\end{equation}
This representation ensures that $\xi(t) \in C^{n}$. 
We introduce the following polynomial approximation of the problem in \eqref{prob: timeopt}.
\begin{problem}
The constrained minimum-control and duration trajectory optimization problem for rigid-body motions on $SE(3)$ is formulated:
\label{prob: mincon}
\begin{subequations}
\begin{align}
\min_{\mathbf{c}, T \in  (0, T_{\text{max}}]} &  \:  \mathcal{J}(\mathbf{c}, T) \quad  \quad  \label{eq:cost_function_new} \\
\text{s.t.} \quad
& \xi^{(k)}(0; \mathbf{c}, T) = \xi_0^{(k)}, \ \forall k = 0, \ldots, s-2, \label{eq:omega_start} \\
& \xi^{(k)}(T; \mathbf{c}, T) = \xi_T^{(k)}, \ \forall k = 0, \ldots, s-2, \label{eq:omega_end} \\
& g_0 \cdot \Phi(\xi(\cdot; \mathbf{c}, T), T) = g_T, \label{eq:rotation_constraint} \\
& \mathcal{G}(\xi(t; \mathbf{c}, T)) \preceq 0, \ \forall t \in [0, T], \label{eq:omega_constraints} 
\end{align}
\end{subequations} 
where \(g_0, g_T \in SE(3)\) denote the initial and terminal poses. $\Phi(\cdot)$ represents the time-ordered integration of the twist trajectory over $[0, T]$, and $\mathcal{G}(\cdot)$ defines additional constraints, such as dynamic feasibility and collision avoidance.
\end{problem}

In the following section, we introduce a learning framework that utilizes this structure to produce trajectories that approximate the optimal solution with high fidelity.

\subsection{Learning Framework}
With boundary conditions consisting of initial and terminal poses $g_0, g_T \in SE(3)$, twists and their derivatives $\xi_0, \xi_T, \dot{\xi}_0, \dot{\xi}_T \in \mathbb{R}^{6}$, and a left-invariant Riemannian metric $W \in \mathbb{S}_{++}^6$, our goal is to predict both a continuous twist trajectory $\xi(t)$ and an execution duration corresponding to the solution of Problem \ref{prob: mincon}.
The trajectory is parameterized as a fixed-degree polynomial that guarantees smoothness and analytic differentiability. 
We employ polynomials of sufficiently high degree $\hat{n}$ to avoid under-parameterization and provide flexibility to satisfy both constraints without artificially clamping higher-order derivatives.

Let $\mathcal{B} = (g_0, g_T, \xi_0, \xi_T, \dot{\xi}_0, \dot{\xi}_T, W)$ collect the boundary data and the left-invariant Riemannian metric.
The learning objective is to construct a smooth map
\begin{equation}
\label{eq: learning}
\mathcal{P}: \mathcal{B} \mapsto \hat{\xi}(\cdot),
\end{equation}
that the trajectory is parameterized by network output.
To reconstruct rigid-body motions, orientations are represented directly in $SE(3)$ for consistent metrics and interpolation properties~\cite{geist2023rotations}. 
We reconstruct the rigid-body trajectory by numerically integrating the
left-trivialized kinematics in \eqref{eq: expmap2} using a
structure-preserving Lie-group integration scheme,
\begin{equation}
\hat{g}_T
=
g_0
\prod_{k=0}^{N-1}
\exp\!\left(
\Omega_k[\hat{\xi}]
\right),
\end{equation}
where $\Omega_k[\hat{\xi}] \in\mathfrak{se}(3)$ denotes the Lie-algebra increment over $[k\Delta t,(k+1)\Delta t]$ computed by the selected numerical integrator.
The ordered product preserves causal composition and provides a computationally efficient and geometrically consistent formulation for $SE(3)$ trajectory learning.

\subsection{Constraint Satisfaction by Construction}
\label{sec:constraint_satisfaction}

Instead of directly outputting $(\mathbf{c}, T)$, we utilize the structure of this problem to enforce a subset of the boundary constraints analytically by construction. 
First, we apply a softplus activation followed by an upper bound and a small offset for numerical stability to satisfy the nonnegative domain of trajectory duration. 
We parameterize duration as
\begin{equation}
\label{eq: time}
T = \min\!\left\{
\log\!\left(1+\exp(\tau)\right)+\epsilon,\,
T_{\max}
\right\}.
\end{equation}
where \( \tau \in \mathbb{R} \) is the raw output of the neural network, and \( \epsilon > 0 \) is a small constant to ensure numerical stability and to avoid zero durations.
This formulation guarantees that \( T \in (\epsilon, T_{\max}] \), keeping the duration strictly positive and bounded. By learning \( \tau \) in an unconstrained space, the network can be trained more effectively while respecting the physical and timing constraints of the trajectory.

We employ coefficient completion to ensure that the generated trajectory strictly satisfies the prescribed higher-order boundary conditions. As an example, consider the minimum-jerk case, for which \(s=3\). The twist trajectory \(\xi(t):[0,T]\rightarrow\mathbb{R}^6\) is represented as a linear combination of polynomial basis functions. The higher-order coefficients are predicted by the network, while the lower-order coefficients are computed analytically from the initial and terminal twist conditions \((\xi_0,\dot{\xi}_0,\xi_T,\dot{\xi}_T)\). These equality constraints are written as
\begin{equation}
\label{eq: linear}
\begin{bmatrix}
1 & 0 & 0 & \cdots & 0 \\
0 & 1 & 0 & \cdots & 0 \\
1 & T & T^2 & \cdots & T^{\hat{n}} \\
0 & 1 & 2T & \cdots & \hat{n}T^{\hat{n}-1}
\end{bmatrix}
\begin{bmatrix}
c_0^\top \\
c_1^\top \\
c_2^\top \\
\vdots \\
c_{\hat{n}}^\top
\end{bmatrix}
=
\begin{bmatrix}
\xi_0^\top \\
\dot{\xi}_0^\top \\
\xi_T^\top \\
\dot{\xi}_T^\top
\end{bmatrix}.
\end{equation}
More generally, the number of free coefficient vectors is
\(n_f=\hat{n}+1-2(s-1)\), and they are collected as
\(\mathbf{c}_f=\big[c_{2s-2},\ldots,c_{\hat{n}}\big]\in\mathbb{R}^{6\times n_f}\).
The coefficient matrix is partitioned as
\(\mathbf{c}=\big[\mathbf{c}_b\;\mathbf{c}_f\big]\), where
\(\mathbf{c}_b\in\mathbb{R}^{6\times(2s-2)}\) contains the boundary-constrained coefficients. The neural network predicts \(\mathbf{c}_f\), while \(\mathbf{c}_b\) is obtained by solving the linear system induced by the boundary conditions.
The output trajectory satisfies \eqref{eq:omega_start} and \eqref{eq:omega_end} by construction, allowing the learning part to focus on minimizing the pose error.
Hence, the mapping in \eqref{eq: learning} can be rewritten as $
\mathcal{P}: \mathcal{B} \mapsto (\mathbf{c}_f, \tau) $,
where the twist trajectory parameters can be recovered with \eqref{eq: time} and \eqref{eq: linear}. For brevity, we denote $\hat{\xi} = \xi(\cdot; \mathbf{c}_f, \tau)$. 
\subsection{Loss Functions}

The training objective is designed to generate trajectories that are physically consistent, energy-time efficient, and compliant with task-related constraints. As an example, we incorporate dynamic feasibility constraints into the problem.
\begin{problem}
The corresponding ideal constrained problem with dynamic feasibility constraints is formulated as
\label{prob: mincon2}
\begin{subequations}
\begin{align}
\min_{\mathbf{c}_f, \tau} & \  \mathcal{J}_m(\hat{\xi}^{(s-1)}) + \alpha_e \|\mathcal{E}(\hat{\xi}^{[2s-1])}) \|^2  +    \alpha_t f(T(\tau)) \quad   \\
\text{s.t.} \quad
& \hat{g_T} = g_T, \\
&\sum_{i = 0}^{s-2}
\phi \left(
\|\hat{\xi}^{(i)}(t)\| -
\lambda_{d,i} \xi^{(i)}_{\max}
\right)^p
= 0  \quad \forall t\in[0,T].
\end{align}
\end{subequations}    
where $\xi_{\max}$ and $\dot\xi_{\max}$ denote the bounds for the velocity and acceleration, \( \lambda_{d,i} \) are scaling factors that adjust the maximum magnitudes, and $p \ge 1$ determines the penalty order. $\mathcal{E}(\cdot)$ is residual of the Euler-Lagrange equation (\ref{eq: euler}), weighted by $\alpha_e$, and \(\phi(\cdot)=\max(0,\cdot)\). 
\end{problem}
In this problem, other constraints are satisfied by construction.
The objective penalizes both the integrated magnitude of higher-order covariant derivatives and Euler-Lagrange residuals.
The latter provides a necessary optimality condition for the problem without additional constraints and serves as a regularization, but does not ensure global optimality.
The residuals are approximated and evaluated in a discretized form.
The time regularization function is chosen as $f(T)=T$.
The overall loss can be written as
\begin{equation}
    \hat{\mathcal{J}} =
    \mathcal{J}_m(\hat{\xi}^{(s-1)})
    + \alpha_e \tfrac{1}{M} \sum_{k=1}^M   \|\mathcal{E}(\hat{\xi}^{[2s-1])}(\tfrac{k}{M}  T))  \|^2 + \alpha_t T(\tau).
\end{equation}
To enforce constraints during training, we adopt an augmented Lagrangian formulation
\begin{equation}
\min \mathcal{L}
=\hat{\mathcal{J}} +\sum_{i \in {\mathrm{pose},\mathrm{dyn}}}
\left( \lambda_i\phi(c_i) + \frac{\rho_i}{2}\phi(c_i)^2
\right)
\end{equation}
where $\lambda_i\in\mathbb{R}_+$ are Lagrange multipliers and $\rho_i>0$ are penalty parameters.
The multipliers are updated online as
\[
\lambda_i
\leftarrow
\Pi_{[0,\lambda_{\max}]}
\!\left(
\lambda_i + \eta_\lambda\,\phi(c_i)
\right),
\]
where $\Pi$ denotes projection onto the feasible interval.
Since higher-order geodesics under a general left-invariant metric on $SE(3)$ do not admit closed-form expressions, we introduce an approximate local metric-weighted geodesic loss to penalize terminal pose error
\begin{equation} c_{\mathrm{pose}} \;=\; \bigl\| \log\left( g_T^{-1} \hat{g_T} \right) \bigr\|_{W}^{2}. 
\end{equation}
The dynamic feasibility constraints are enforced at discretized points.
The corresponding penalty is defined as
\begin{equation}
c_{\mathrm{dyn}}
= \tfrac{1}{M} \sum_{k=1}^{M}
\sum_{i = 0}^{s-2}
\phi \left(
\bigl \| \hat{\xi}^{(i)}(\tfrac{k}{M}  T) \bigr \|-
\lambda_{d,i} \xi^{(i)}_{\max}
\right)^p
\end{equation}
This formulation softly enforces limits on angular velocity and higher-order derivatives by penalizing violations at sampled points along the trajectory.

In addition to boundary conditions and dynamic feasibility, many practical robotic applications impose task-specific or hardware-related requirements that must be incorporated during trajectory generation.
Examples include limits on actuator saturation, avoidance of singular configurations, and field-of-view constraints for onboard sensors.
Such requirements can be incorporated within the same framework as additional constraint terms.
Soft penalties allow the optimization to remain differentiable and numerically stable during training, while still encouraging satisfaction of these constraints whenever feasible.
Further examples for underactuated systems are discussed in Section~\ref{sec: applications}.

\subsection{Metric-Conditioned Trajectory Learning}
\label{sec: metric_hyper}

To enable adaptation across different left-invariant Riemannian metrics, we introduce a metric-conditioned parameterization that modulates the trajectory network based on the metric tensor. 
For numerical stability, we regularize the metric as
\begin{equation}
\widetilde{W} = \tfrac{1}{2}(W+W^\top)+\delta I_6,
\qquad \delta>0,
\end{equation}
and apply a structured factorization $S(\cdot)$ (e.g., Cholesky) to obtain a compact representation
\begin{equation}
u = \operatorname{vech}\!\big(S(\widetilde{W})\big) \in \mathbb{R}^{21}.
\end{equation}
The encoded metric $u$ is mapped to a latent embedding $z = \Phi_m(u)$, which conditions each network block via a modulation operator $\Psi_i$. 
The $i$-th layer is given by
\begin{equation}
h_i = \varphi \left( \mathcal{N}_i(\Psi_i(h_{i-1}; z)) + r_i(h_{i-1}) \right),
\end{equation}
where $\mathcal{N}_i(\cdot)$ denotes normalization and $r_i(\cdot)$ is a residual connection.
This generates a metric-dependent trajectory mapping through the modulation operators. 
The final trajectory is constructed as $(\mathbf{c}_f,\tau)=\mathcal{P}_{\theta}(\mathcal{B})$ and generalizes across heterogeneous and anisotropic metrics.

\section{Results}
\label{sec:evaluations}
We conduct comprehensive evaluations of the proposed framework across multiple experimental settings to highlight the following aspects:
(1) \textbf{Trajectory quality:} Produces smooth, near-optimal trajectories that minimize cost functionals under arbitrary state conditions and left-invariant Riemannian metrics.  
(2) \textbf{Ablation study:} Ablation studies show that the energy formulation and dynamics constraints are essential. Model variants consistently underperform, confirming alignment with the underlying geometry.
(3) \textbf{Benchmarks:} Provide improved reliability, boundary accuracy, dynamic feasibility, and computational efficiency compared with classical and neural BVP
solvers.
We aim to demonstrate that the proposed learning framework is consistent with left-invariant Riemannian geometry while improving trajectory quality and computational efficiency.
\subsection{Dataset Generation}

We construct a synthetic dataset of motion samples for training and evaluating our $SE(3)$ dynamics model. 
The initial and terminal poses are independently drawn by uniformly sampling rotations from $SO(3)$ and translations on the specified mode. 
Twists and twist derivatives are sampled from uniform distributions scaled by prescribed factors. 
Boundary conditions are imposed by clamping variables according to the assigned phase, ensuring that the sampled states are physically consistent. 
We keep both forward and time-reversed boundary-condition pairs because reversing a trajectory changes the boundary derivatives and provides a distinct conditioned input for the learning model.

\subsubsection{Boundary conditions (motion phase)} 
We categorize trajectories into three phases that reflect increasing kinematic freedom and dynamic complexity. (1) \emph{Rest motion (R$\leftrightarrow$R):} both endpoints are fully at rest, with zero twists and accelerations. This represents the simplest boundary-value case, corresponding to motions that start and end at rest.
(2) \emph{Semi-rest motion (R$\leftrightarrow$M):} one boundary state is in motion while the other is static, representing transitions that merge into or replanning from a non-rest state.
(3) \emph{Non-rest motion (M$\leftrightarrow$M):} both endpoints are in motion, allowing nonzero velocities and accelerations at the boundaries. This case represents fully dynamic transitions between moving states.
The three motion categories reflect the boundary-value conditions encountered in planning and control, from static to fully dynamic motion (for replanning), while avoiding unnecessary subdivision of intermediate cases such as semi-rest. 
Throughout all phases, positions remain continuous, velocities vary smoothly without discontinuities, and accelerations may change instantaneously through control inputs.

\subsubsection{Metric difficulty} 
We incorporate different metrics ranging from isotropic to anisotropic and frame-dependent. (1) $W_0$: isotropic metric with scalar weights, identical across all rotational and translational directions. (2) $W_H$: rigid-body inertia metric that combines principal moments of inertia and isotropic mass. (3) $W_B$: anisotropic metric with axis-dependent translational weights (block-diagonal structure). (4) $W_C$: isotropic mass metric in a rotated and displaced frame, with coupling between rotation and translation. (5) $W_{CB}$: anisotropic block-diagonal metric expressed in a rotated and displaced frame.
Training begins with simple isotropic metrics and gradually progresses toward anisotropic, frame-dependent, and unstructured metrics, which present the most challenging learning conditions. 

\begin{figure}[!th]
\centering
\includegraphics[width=0.98\columnwidth]{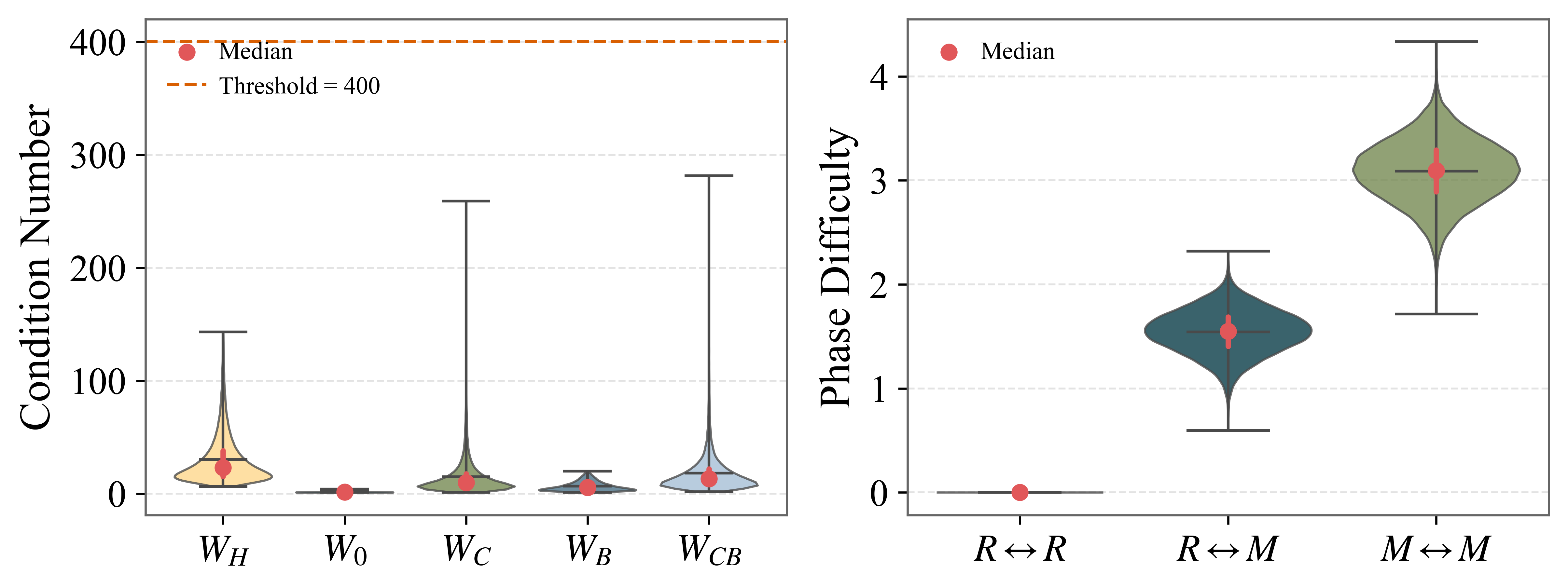}
\vspace{-0.2cm}
      \caption{Evaluation of different metrics difficulty in condition number (left) and motion phase difficulty in boundary-rate score(right) .}
      \label{fig: diff}
     \vspace{-0.2cm}
\end{figure}

We further qualify the difficulty level of samples for different metrics and motion phases.
For each metric, we compute the spectral condition number of the metric matrix. 
To avoid unstable samples during the training, we set a threshold to filter ill-conditioned samples. 
For motion phase difficulty, we compute a boundary-rate difficulty score defined as the sum of two normalized terms: the endpoint velocity magnitude divided by its allowable maximum, and the endpoint acceleration magnitude divided by its allowable maximum. 
The velocity and acceleration used in this score correspond to the larger of the two endpoints in each case.
The distributions in Fig.~\ref{fig: diff}
characterize metric conditioning and phase-wise motion difficulty in the synthetic dataset.

\begin{figure*}[!t]
 \includegraphics[width=2.0\columnwidth]{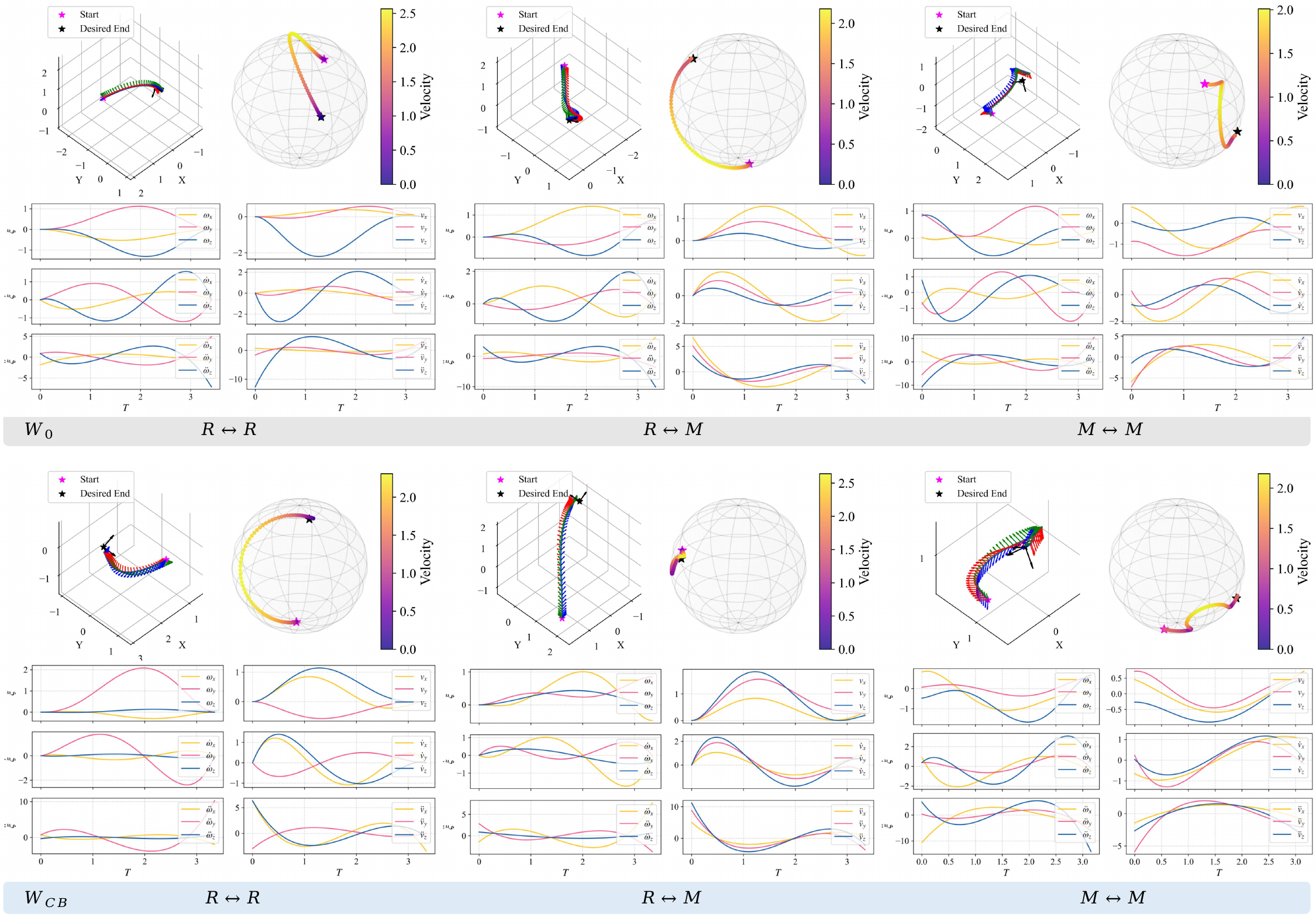}
    \caption{Two representative metric classes ($W_0$ and $W_{CB}$) with trajectories from all motion phases generated by the proposed framework. 
    For each case, we show the $\mathbb{R}^3$ trajectory with body axes (left) and the corresponding $SO(3)$ trajectory on the unit sphere~\cite{doi:10.1137/17M1129416}, where color encodes velocity magnitude (right). The bottom plot of each case shows twist profiles over time, including higher-order derivatives, with colors indicating axes.
    }
    \vspace{-0.3cm}
    \label{fig: trajectory}
\end{figure*}

For each metric class and motion phase, we generate an independent dataset and store it separately. 
To improve model robustness, small perturbations are applied to poses, velocities, accelerations, and timing, ensuring stability under minor variations during inference.
This yields a two-dimensional curriculum: models can first be trained on easy cases (R$\to$R with $W_0$) and then gradually be exposed to harder boundary conditions and increasingly complex metrics. 
The resulting curriculum provides balanced coverage of the motion space, avoids degenerate or inconsistent trajectories, and supports smooth generalization across diverse SE(3) dynamics.

\subsection{Implementation Details}

The trajectory network is a metric-conditioned multilayer perceptron with four hidden layers of width 384, a shared backbone, and dual output heads for polynomial twist coefficients and the optimal duration \(T_{\mathrm{opt}}\). 
To incorporate metric dependence, we employ two conditioning networks: FiLM-based affine modulation and low-rank metric-conditioned adaptation, both driven by a Cholesky-based embedding of \(W\in\mathbb{S}_{++}^{6}\). 
The backbone is further stabilized by lightweight architectural refinements, including boundary-attention and post-trunk residual refinement. 
$\phi(\cdot)$ is implemented via $\mathrm{ReLU}$, and we use $p=2$.
We set the polynomial degree up to \(n=12\), as higher-order parameterizations were observed to be less numerically stable~\cite{richter2016polynomial}.

Training is conducted for 5000 epochs under a hybrid curriculum with a batch size of 512. 
For ablation studies and architectural variations, all models are trained for 1000 epochs.
The Euler-Lagrange and dynamic feasibility losses are evaluated at \(N \in [50,100]\) collocation points along each trajectory. 
We set the maximum duration of $T_{\max} = 10\,\mathrm{s}$, with $\epsilon = 0.01$, to encourage convergence to feasible and efficient motions. 
All models are trained and evaluated on NVIDIA L40S GPUs.
Other comparisons are conducted on an i7-11800H CPU.
For visualization of orientations, \(SO(3)\) trajectories are projected onto the unit sphere following~\cite{doi:10.1137/17M1129416}, by tracking the evolution of \(R(t)v\), where \(v = \tfrac{1}{\sqrt{3}}[1,1,1]^T\). 

\subsection{Numerical Evaluation}

\subsubsection{Trajectory Performance}
We first present a representative example of a generated $SE(3)$ trajectory, shown in Fig.~\ref{fig: trajectory}.  
The framework produces smooth, continuous motions that connect the initial and terminal states. The terminal rotation error remains within a small margin in the goal region, confirming that the boundary conditions are satisfied. The pose loss is on the same order as typical trajectory tracking errors and is therefore negligible in practice for robotic applications.  

\begin{figure}[!th]
  \centering
  \includegraphics[width=1\columnwidth]{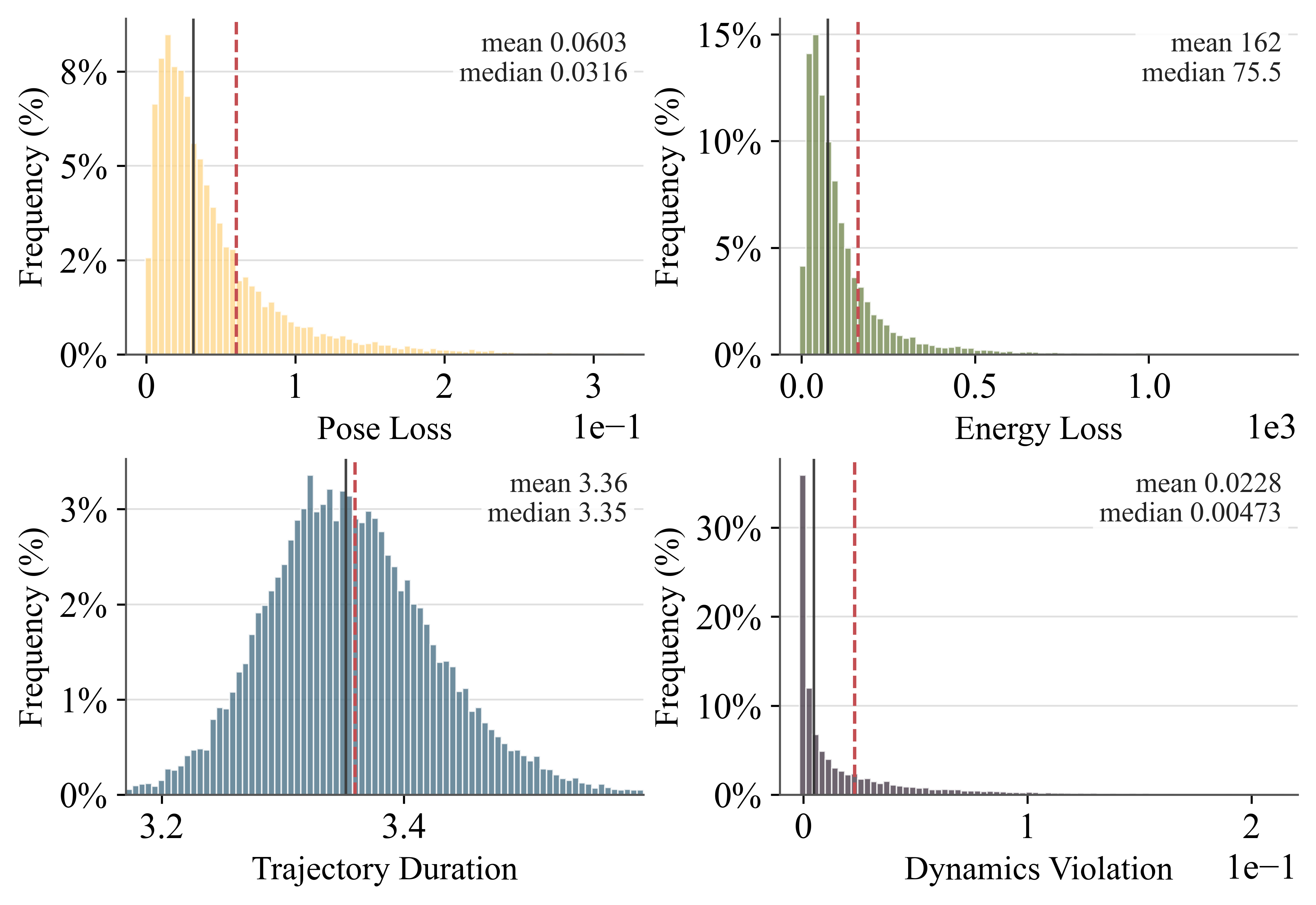}
  \vspace{-0.3cm}
  \caption{Distribution of evaluation metrics.
Histograms show the distribution of sample-wise values for pose loss, energy loss, trajectory time duration, and dynamic violation. Red dashed lines denote the mean of each metric.}
  \label{fig: numerical}
  \vspace{-0.3cm}
\end{figure}

Furthermore, we show aggregated results 
in the split test dataset (i.e., $15{,}000$ trajectories in total) with one loss recorded for each trajectory.  
As summarized in Fig.~\ref{fig: numerical}, we evaluate different loss terms and dynamic violation.  
On average, the framework achieves an $SE(3)$ error in the order of 0.1 while keeping velocity and acceleration limit violations around 0.02, respectively.  
Residual smoothness loss remains due to the finite-degree polynomial representation, reflecting the approximation limits of the parameterization.
The average inference plus integration time for generating the batch is within \(1.0\,\mathrm{ms}\) on average, supporting real-time motion primitive
generation.

\subsubsection{Robustness Analysis}

\begin{figure}[!th]
  \centering
  \includegraphics[width=1\columnwidth]{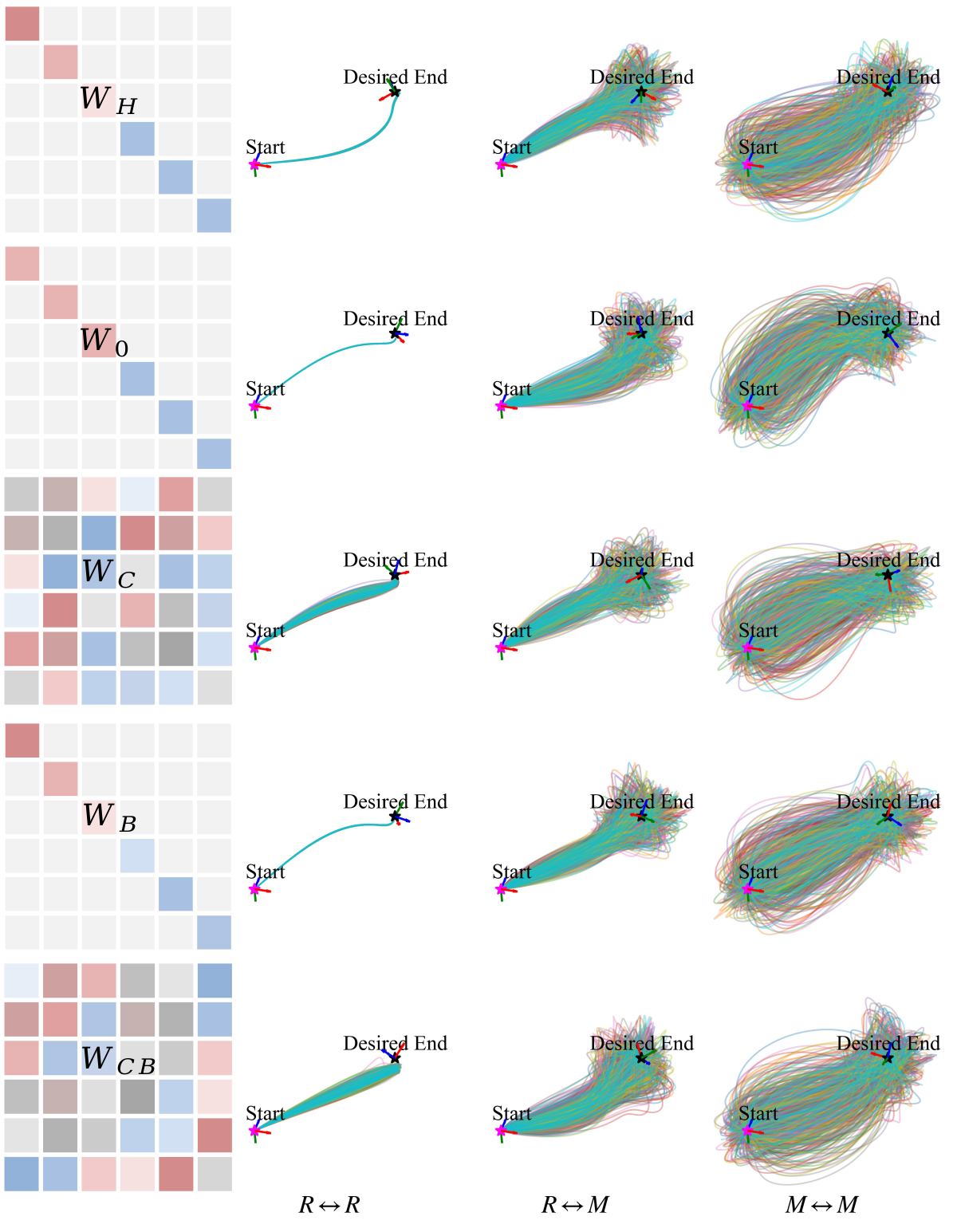}
  \vspace{-0.2cm}
  \caption{Trajectories between fixed random initial and terminal poses under varying metric and motion conditions. Each row corresponds to a different class of metrics (five in total). Within each row, columns (from left to right) represent rest motions, semi-rest motions, and non-rest motions, with randomly sampled boundary twists and their derivatives.
  }
  \label{fig: diverse}
  \vspace{-0.2cm}
\end{figure}

We evaluate the ability of the framework to generate diverse trajectories
under identical endpoint poses but varying higher-order conditions.
Fig.~\ref{fig: diverse} illustrates one batch of trajectories obtained in a single forward pass, each with different randomly sampled boundary conditions and metrics.

The predicted curves represent feasible motion primitives that remain
smooth and dynamically consistent with the specified boundary conditions.
In the R$\leftrightarrow$R case, the metric is the only varying factor across rows.
The observed behavior is consistent with the metric structure in the
variational formulation. When the metric does not introduce coupling and the boundary conditions are at rest, the Euler-Lagrange equations decouple
across motion components, so the learned trajectories concentrate near
the corresponding optimal solution. 
When additional body
frames with rigid transformations are introduced (\(W_C, W_{CB}\)),
rotational and translational components become coupled in the
Euler-Lagrange equations, and different transformations lead to distinct
optimal trajectories.

\subsection{Ablation and Model Variants}

We conduct a comprehensive study to quantify the impact of loss components and network design. 
All variants are trained with the same hyperparameters and datasets for a fair comparison. 
For the loss ablation experiments, all variants use FiLM conditioning for metric embedding.
Performance is evaluated in terms of trajectory smoothness (energy and Euler-Lagrange residual), boundary accuracy (velocity/acceleration and pose errors), trajectory duration, and dynamic feasibility. Results are summarized in Table~\ref{tab: ablation1} and~\ref{tab: ablation2}.

\begin{table}[!th]
\centering
\renewcommand\arraystretch{1.2}
\setlength{\tabcolsep}{2.2pt}
\caption{Comparison of loss component ablation (mean $\pm$ std).}
\vspace{-0.1cm}
\label{tab: ablation1}
\begin{tabular}{lP{1.2cm}P{1.2cm}P{0.9cm}P{0.88cm}P{0.8cm}P{1.35cm}}
\toprule
\textbf{Model }& \multicolumn{2}{c}{\textbf{Smoothness $\downarrow$}} & \multicolumn{2}{c}{\textbf{Boundary Error $\downarrow$}}& \multirow{2}{*}{\begin{tabular}{l}\textbf{Dur.}  \\
\textbf{$T$(s)}\end{tabular}} & \multirow{2}{*}{\begin{tabular}{l}\textbf{Dynamics} \\
 \textbf{Violation $\downarrow$}\end{tabular}} \\
 \cline{2-3}\cline{4-5}
& \textbf{Energy} & \textbf{E.L.} & \textbf{Vel/Acc} & \textbf{Pose} & & \\
\hline
Ours & \textbf{210.050 (177.016)} & \textbf{1.24e+03 (1.32e+03)} & 0.000 (0.000) & \textbf{0.113 (0.055)} & 3.161 (0.050) & \textbf{0.021 (0.012)} \\
w/o $\mathcal{J}_{m}$ & 339.207 (327.602) & 1.97e+03 (2.33e+03) & 0.000 (0.000) & 0.144 (0.068) & 3.103 (0.025) & 0.030 (0.014) \\
w/o $\mathcal{E}$ & 227.798 (201.156) & 1.37e+03 (1.47e+03) & 0.000 (0.000) & 0.135 (0.065) & 3.089 (0.037) & 0.022 (0.012) \\
w/o $T$ & 248.516 (240.617) & 1.57e+03 (1.88e+03) & 0.000 (0.000) & 0.143 (0.065) & 3.043 (0.031) & 0.025 (0.013) \\
w/o $c_{\mathrm{dyn}}$ & 395.606 (260.263) & 2.52e+03 (2.25e+03) & 0.000 (0.000) & 0.142 (0.064) & \textbf{2.553 (0.080)} & 0.130 (0.063) \\
\bottomrule
\end{tabular}
\vspace{-0.2cm}
\end{table}

\paragraph{Loss Component Ablation}

The full model with all losses achieves the best overall performance. The ablation results demonstrate that the smoothness and dynamics-related losses are essential for maintaining trajectory quality and feasibility.
Removing $\mathcal{J}_{m}$ or dynamics constraint loss $c_{\mathrm{dyn}}$ significantly degrades smoothness.
Removing the Euler-Lagrange residual $\mathcal{E}$ or the time regularization term $T$ results in more moderate degradation. This behavior is expected: once boundary and other constraints are enforced, trajectories may deviate from the exact Euler-Lagrange solution but remain feasible. 
The weight associated with $T$ is intentionally small since $T$ is implicitly regulated by other losses such as the dynamics constraints and energy.
In practice, trajectories balance energy optimality and duration via loss weight adjustment.
The most significant degradation occurs when the dynamics constraint $c_{\mathrm{dyn}}$ is removed. 
Although this variant achieves the shortest trajectory duration ($2.553$~s), the resulting trajectories have larger violations of dynamic limits.

\begin{table}[!ht]
\centering
\vspace{0.2cm}
\renewcommand\arraystretch{1.2}
\setlength{\tabcolsep}{2.2pt}
\caption{Comparison of other model variants (mean $\pm$ std).}
\label{tab: ablation2}
\begin{tabular}{P{1.15cm}P{1.2cm}P{1.2cm}P{0.9cm}P{0.88cm}P{0.8cm}P{1.35cm}}
\toprule
\textbf{Model }& \multicolumn{2}{c}{\textbf{Smoothness $\downarrow$}} & \multicolumn{2}{c}{\textbf{Boundary Error $\downarrow$}}& \multirow{2}{*}{\begin{tabular}{l}\textbf{Dur.}  \\
\textbf{$T$(s)}\end{tabular}} & \multirow{2}{*}{\begin{tabular}{l}\textbf{Dynamics} \\
 \textbf{Violation $\downarrow$}\end{tabular}} \\
 \cline{2-3}\cline{4-5}
& \textbf{Energy} & \textbf{E.L.} & \textbf{Vel/Acc} & \textbf{Pose} & & \\
\hline
Ours (FiLM) & 210.050 (177.016) & 1.24e+03 (1.32e+03) & \textbf{0.000 (0.000)} & \textbf{0.113 (0.055)} & 3.161 (0.050) & \textbf{0.021 (0.012)} \\
Ours (low-rank) & 224.528 (187.066) & 1.34e+03 (1.40e+03) & 0.000 (0.000) & 0.135 (0.074) & 3.103 (0.035) & 0.022 (0.012) \\
Soft (FiLM) & 157.534 (159.031) & 701.882 (863.820) & 0.350 (0.271) & 0.161 (0.061) & 3.819 (0.052) & \textbf{0.006 (0.003)} \\
Soft (low-rank) & \textbf{143.572 (136.246)} & \textbf{533.485 (608.702)} & 0.312 (0.273) & \textbf{0.086 (0.034)} & 4.189 (0.063) & \textbf{0.006 (0.003)} \\
Start (FiLM) & 198.968 (201.398) & 1.24e+03 (1.63e+03) & 0.210 (0.091) & 0.120 (0.057) & 3.149 (0.223) & 0.030 (0.029) \\
Start (low-rank) & 225.404 (225.974) & 1.27e+03 (1.64e+03) & 0.185 (0.076) & 0.158 (0.080) & 3.279 (0.281) & 0.029 (0.028) \\
\bottomrule
\end{tabular}
\vspace{-0.2cm}
\end{table}

\paragraph{Architectural Variants}
We further analyze the effects of the coefficient completion module and the metric embeddings.
Two alternative boundary enforcement strategies (\textit{Soft} and \textit{Start}) are compared. The \textit{Soft} variant directly predicts all polynomial coefficients without explicit boundary enforcement, achieving the lowest smoothness metrics but suffering from poor boundary accuracy, with velocity and acceleration errors between $0.31$ and $0.35$. 
The \textit{Start} variant partially enforces boundary constraints by fixing the initial twist and acceleration coefficients, which reduces these errors but restricts the optimization space while still failing to guarantee endpoint feasibility. The proposed analytic completion module solves a closed-form linear system to enforce boundary conditions exactly.
Although the trajectory energy is slightly higher than the unconstrained \textit{Soft} variant, it achieves strict boundary satisfaction while maintaining dynamic consistency.

The comparison of metric embeddings shows that, when combined with the analytic completion module, both FiLM and low-rank conditioning achieve similar performance. FiLM provides a slight improvement, while the overall behavior remains largely governed by strict boundary enforcement.
The low-rank hypernetwork provides a slight improvement in smoothness under the Soft configuration, consistent with its higher representational capacity.

Overall, the results demonstrate the trade-off between trajectory optimality and constraint satisfaction. 
The unconstrained formulation achieves lower smoothness metrics but suffers from boundary errors, whereas the analytic completion guarantees exact boundary feasibility while maintaining competitive smoothness and runtime performance.

\begin{table*}[t]
\caption{Benchmark comparison across different metrics (mean $\pm$ std).}
\centering
\renewcommand\arraystretch{1.2}
\resizebox{2\columnwidth}{!}{%
\begin{tabular}{ll|cllllll}
\toprule
\textbf{Metric} & \textbf{Method} &
\multirow{2}{*}{\begin{tabular}{c}\textbf{Success}\\\textbf{Rate (\%)} $\uparrow$\end{tabular}} &
\multicolumn{2}{c}{\textbf{Smoothness $\downarrow$}} &
\multicolumn{2}{c}{\textbf{Boundary Error $\downarrow$}} &
\multirow{2}{*}{\begin{tabular}{c}\textbf{Dynamics}\\\textbf{Violation} $\downarrow$\end{tabular}} &
\textbf{Solve Time (s) $\downarrow$} \\
\cline{4-5}\cline{6-7}
& & & \textbf{Energy} & \textbf{E.L.} & \textbf{Vel/Acc} & \textbf{Pose} & & \\
\midrule

\multirow{3}{*}{$W_0$}
 & Ours       & 100.0 & 79.245 (47.698) & 317.540 (216.655) & 0.000 (0.000) & 0.051 (0.041) & 0.032 (0.050) & 0.001 (0.000) \\
 & BVP-NN       & 100.0 & 1934.492 (2576.451) & 17.146 (17.799) & 0.040 (0.043) & 0.002 (0.005) & 0.859 (0.992) & 0.000 (0.000) \\
 & BVP-Opt & 70.0 & 1.236 (1.448) & 0.222 (0.182) & 1.159 (0.451) & 0.795 (0.742) & 0.000 (0.000) & 1000.419 (313.172) \\
\hline
\multirow{3}{*}{$W_B$}
 & Ours      & 100.0 & 170.024 (233.788) & 1334.677 (1887.795) & 0.000 (0.000) & 0.046 (0.036) & 0.017 (0.036) & 0.001 (0.000) \\
 & BVP-NN       & 100.0 & 2369.649 (4510.718) & 132.601 (186.003) & 0.277 (0.531) & 0.072 (0.187) & 0.346 (0.371) & 0.000 (0.000) \\
 & BVP-Opt & 83.3 & 2.434 (3.551) & 0.246 (0.181) & 1.199 (0.526) & 0.924 (1.251) & 0.000 (0.000) & 1010.509 (323.467) \\
\hline
\multirow{3}{*}{$W_C$}
 & Ours       & 100.0 & 124.315 (127.659) & 590.044 (505.491) & 0.000 (0.000) & 0.057 (0.049) & 0.029 (0.070) & 0.001 (0.000) \\
 & BVP-NN     & 100.0 & 2542.547 (3327.644) & 23.558 (26.127) & 0.034 (0.021) & 0.002 (0.006) & 0.649 (0.929) & 0.000 (0.000) \\
 & BVP-Opt & 76.7 & 2.360 (3.175) & 0.214 (0.170) & 1.215 (0.564) & 0.847 (0.781) & 0.000 (0.000) & 931.545 (377.360) \\
\hline
\multirow{3}{*}{$W_{CB}$}
 & Ours       & 100.0 & 184.807 (181.421) & 1560.897 (1805.933) & 0.000 (0.000) & 0.037 (0.044) & 0.010 (0.016) & 0.001 (0.000) \\
 & BVP-NN      & 100.0 & 1652.918 (2281.722) & 162.456 (165.743) & 0.226 (0.285) & 0.091 (0.191) & 0.375 (0.441) & 0.000 (0.000) \\
 & BVP-Opt & 73.3 & 3.490 (5.288) & 0.221 (0.125) & 1.214 (0.472) & 0.766 (0.675) & 0.000 (0.000) & 1060.820 (361.206) \\
\hline
\multirow{3}{*}{$W_H$}
 & Ours       & 100.0 & 61.502 (55.327) & 312.191 (218.570) & 0.000 (0.000) & 0.022 (0.022) & 0.022 (0.031) & 0.001 (0.000) \\
 & BVP-NN       & 100.0 & 1986.448 (2512.133) & 14.120 (12.342) & 0.032 (0.008) & 0.001 (0.001) & 0.835 (0.910) & 0.000 (0.000) \\
 & BVP-Opt & 90.0 & 1.264 (2.492) & 0.282 (0.193) & 1.228 (0.443) & 0.531 (0.463) & 0.000 (0.000) & 798.603 (314.565) \\
\bottomrule
\end{tabular}}
\label{tab: benchmark}
\vspace{-0.2cm}
\end{table*}

\subsection{Benchmark Comparison}

Closed-form solutions are generally unavailable for higher-order trajectory generation under left-invariant Riemannian metrics. 
We therefore benchmark the proposed method against two representative baselines: a collocation-based boundary value problem solver, denoted BVP-Opt, which directly solves the Euler–Lagrange equations using SciPy~\cite{2020SciPy-NMeth}, and a neural boundary value problem solver, denoted BVP-NN, implemented as a per-instance physics-informed neural network~\cite{RAISSI2019686} trained by minimizing the Euler–Lagrange residual.
Optimizing the trajectory duration jointly with the state significantly increases the difficulty for the baseline methods. 
To ensure a fair comparison, the trajectory time produced by the proposed method is fixed and used by the baselines. 
For the BVP solver, the maximum number of iterations is set to $100$, and runs that fail to converge within this limit are classified as failures.
All methods use the same discretization resolution and are evaluated on five metric families with 30 randomized boundary conditions per metric.

As shown in Table~\ref{tab: benchmark}, the proposed method achieves high-quality trajectories across all metrics while maintaining negligible pose errors and dynamics violations. 
The BVP solver succeeds in only $70\%$--$90\%$ of cases and often produces large boundary errors, although it can achieve low energy when convergence occurs. 
The neural BVP also achieves good boundary satisfaction, but has substantially higher dynamics violations. 
In our experiments, a separate per-instance neural BVP is trained for each boundary condition; learning a single conditioned BVP that generalizes across boundary conditions often degrades solution quality.
Our experiments indicate that the BVP solver rarely converges to satisfactory solutions within the allotted time.
Overall, the proposed method provides the most reliable and computationally efficient solution while maintaining strong boundary accuracy and dynamic feasibility.

\section{Applications}
\label{sec: applications}

\subsection{SE(3) Motion and Trajectory Generation}

The proposed framework generates smooth reference trajectories in \(SE(3)\).
For a fully actuated system, any sufficiently smooth trajectory of position
and orientation, together with their time derivatives, is theoretically
trackable provided the required control inputs remain within actuator limits.

\subsubsection{Primitive Generation}

\begin{figure}[!th]
\includegraphics[width=0.95\columnwidth]{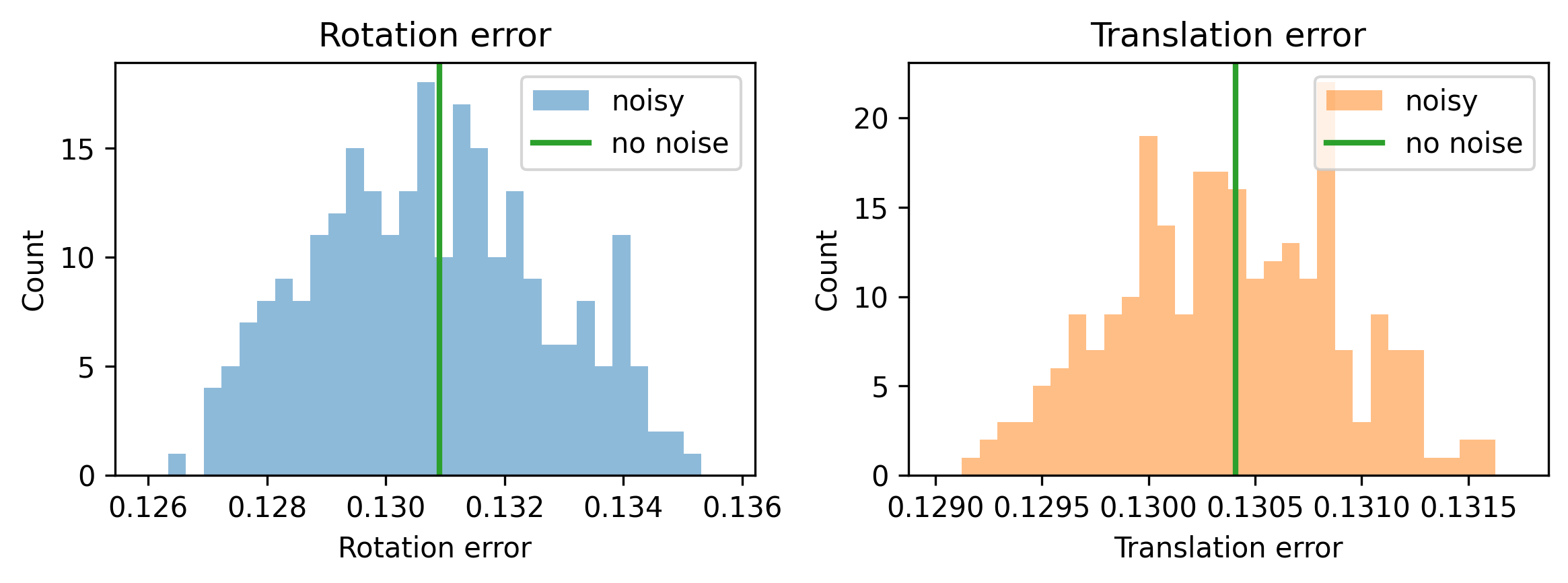}
    \vspace{-0.1cm}
    \caption{Rotation and translation errors with perturbed end poses.}
    \vspace{-0.2cm}
    \label{fig: single_motion_errors}
\end{figure}

Given a fixed initial condition, we generate candidate trajectories in a single forward inference by sampling perturbed terminal boundary conditions. Each candidate produces a terminal pose \(g_T^{(i)}\), and the final trajectory is selected by minimizing the terminal pose error. Fig.~\ref{fig: single_motion_errors} shows representative rotation and translation errors from one trial. This procedure enables efficient parallel exploration of boundary conditions.

The above formulation generates a single primitive with a fixed terminal pose. In many applications, the goal need not be restricted to a single point, and reaching a feasible terminal region is sufficient. For a fixed start and end pose pair \((g_0, g_T)\), the method samples boundary states \((\xi_0,\xi_T,\dot{\xi}_0,\dot{\xi}_T)^{(i)}\), \(i=1,\dots,N\), to construct a a motion primitive library.
More advanced primitive library construction, such as ~\cite{9561840,11015263}, can be integrated within this framework.

\subsubsection{Piecewise Trajectory Synthesis}

To traverse multiple waypoints with higher-order conditions, the single-segment primitive generation can be applied sequentially to each pair of boundary conditions.
The terminal twist state of segment $k\!-\!1$ is reused as the initial twist state of segment $k$:
\begin{equation}
    \xi_k(0) \leftarrow \xi_{k-1}(T_{k-1}), \qquad
\dot{\xi}_k(0) \leftarrow \dot{\xi}_{k-1}(T_{k-1}).
\end{equation}
which enforces $C^1$ continuity in twist at intermediate waypoints. 
The whole trajectory is obtained by concatenating all segments. 
Fig.~\ref{fig: fig1}(a) illustrates an example instance, with $SO(3)$ trajectories in different views (left) and the corresponding $SE(3)$ trajectory (right).

\subsubsection{Replanning under Non-Rest Boundary Conditions}

\begin{figure}[!ht]
\includegraphics[width=0.95\columnwidth]{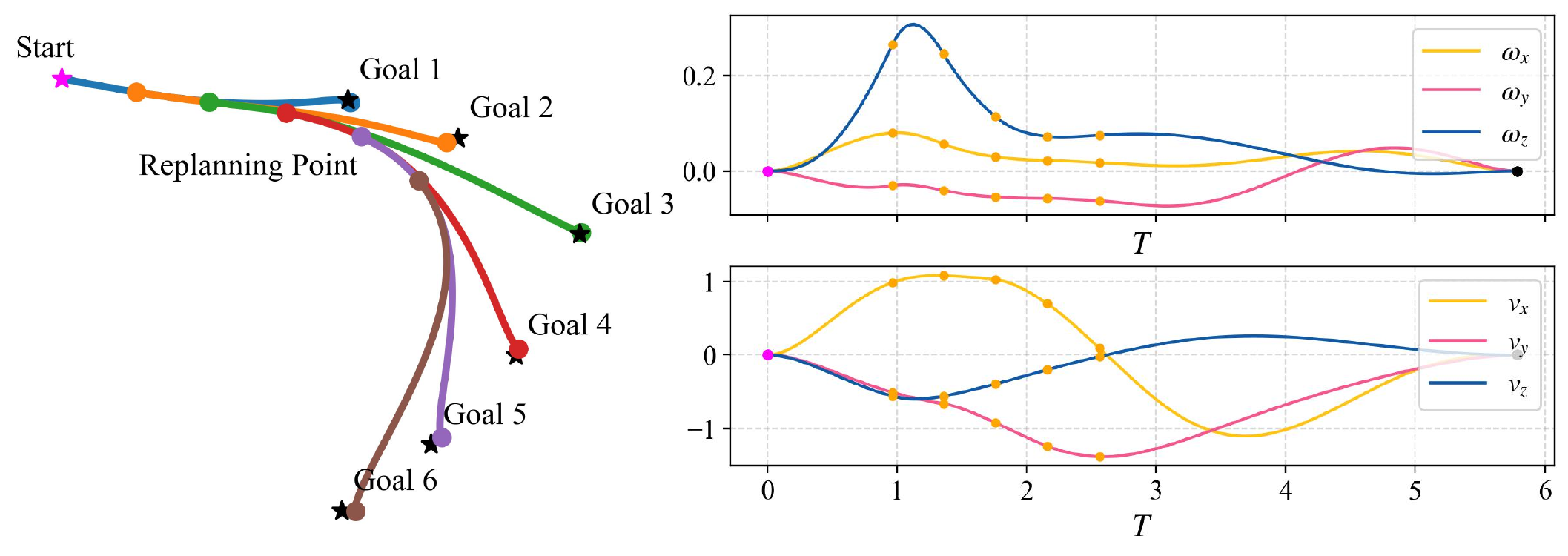}
    \vspace{-0.1cm}
    \caption{Replanning under non-rest motions. The left shows trajectories colored by replanning segments, and the right shows the twist profiles. }
    \vspace{-0.4cm}
    \label{fig: replanning}
\end{figure}

The framework also supports replanning under non-rest boundary conditions (nonzero velocity and acceleration), which is useful for tasks such as landing and dynamic obstacle avoidance.  
Fig.~\ref{fig: replanning} shows an example where an initial trajectory is generated from the start state to the first goal using the proposed framework. 
Replanning is triggered at \(30\%\) of the trajectory and generates a new trajectory toward an updated goal. 
This process continues with additional replanning points and goals.
In addition to primitive libraries, the proposed method allows trajectory generation from arbitrary states during execution, enabling efficient online planning. 
In this work, we demonstrate minimum-jerk trajectories with $\xi \in C^1$ at replanning points.
Higher-order continuity can be incorporated within the same framework by extending the network inputs to include additional boundary derivatives.

\subsection{Model Adaptation for Quadrotor Deployment}

For underactuated platforms, generated $SE(3)$ trajectories must satisfy specific dynamic constraints. While the proposed framework produces metric-consistent geometric trajectories on $SE(3)$, robot feasibility can be incorporated through a modular constraint layer. 
We use a quadrotor as an example and leverage differential flatness to impose thrust-axis alignment and actuator limits during training.

We adopt a two-stage fine-tuning strategy. In the first stage, the pretrained $SE(3)$ generator is adapted to a quadrotor-feasible distribution. 
The dataset is generated using the quadrotor dynamics by sampling flat outputs and reconstructing the corresponding twist and its derivatives, together with a fixed quadrotor kinetic metric $W_{\mathrm{quad}} = \mathrm{diag}(J, m I_3)$.
The model is then fine-tuned on this quadrotor-constrained dataset, initialized from the pretrained $SE(3)$ generator checkpoints.

In the second stage, feasibility constraints are explicitly enforced during training. 
From the reconstructed trajectory, the desired force is $ f^{\mathrm{des}} = m(a + g \mathbf{e}_3)$, with $\mathbf{e}_3=[0,0,1]^T$. 
Since a quadrotor generates thrust along its body $z$-axis, we project the thrust force onto the thrust direction and minimize the mismatch to enforce actuation consistency.
Additional constraints for quadrotor dynamics are encoded on angular rate, tilt angle, thrust limits, and control moments.
Constraint violations are penalized using the same formulation as in the main framework. 
To maintain consistency with the pretrained model, an anchor regularization term encourages the updated trajectories to remain close to those generated in the first stage.
We demonstrate the fine-tuned trajectories using Crazyflie 2.1 Brushless with RotorPy simulations~\cite{folk2023rotorpy} and deploy them in real-world flight tests.
Fig.~\ref{fig: fig1}(b) illustrates six trajectories with 2-6 waypoints, where intermediate states are randomly sampled. This formulation can be naturally extended to waypoint-constrained trajectory optimization frameworks, such as those based on endpoint derivatives or Hermite splines~\cite{richter2016polynomial, 11474851}.

\section{Discussion and Conclusion}

\subsubsection*{Conclusion}
In this work, we introduce a learning-based framework for generating smooth trajectories on $SE(3)$ under general left-invariant Riemannian metrics. 
The polynomial twist parameterization, together with analytic coefficient completion, ensures that the generated trajectories satisfy boundary derivatives by construction while retaining sufficient flexibility for optimization via learning.
The experimental results indicate that incorporating geometric structures significantly improves the quality and stability of the learned trajectories. 
The metric-conditioned architecture further enables adaptation across different metric structures, allowing a single model to generalize over a family of physically meaningful metrics.
As trajectory parameters are predicted in millisecond-scale inference time, the framework can support applications that require rapid replanning or real-time motion-primitive generation.

\subsubsection*{Limitations}
Despite these advantages, several theoretical limitations remain. 
The learned trajectories approximate the solution of the variational optimization problem and do not guarantee exact optimality. 
The accuracy of the approximation depends on the polynomial representation and the coverage of the training distribution. 
When boundary conditions or metric structures fall outside the training distribution, the quality of the generated trajectories may degrade, which is inherent to data-driven approximation methods.
The current formulation focuses on smoothness and dynamic feasibility constraints. 
Other constraints, such as obstacle avoidance, are not explicitly incorporated and evaluated.
In addition, although the metric-conditioned architecture enables generalization across different left-invariant metrics, the approach assumes that the metric remains constant along the trajectory. 
Extensions to configuration-dependent metrics or systems with more complex dynamics would require additional modeling and may introduce further computational challenges.

\subsubsection*{Extension}
The proposed framework is not limited to the applications considered in this work and can be extended to a broader range of robotic platforms and task settings. 
It provides smooth, dynamically consistent trajectory initializations that can be further refined by feedback controllers or trajectory optimization methods.
It can also serve as an efficient initialization for other trajectory optimization methods, reducing computational burden in real-time settings.
Additional constraints, such as obstacle avoidance, can be incorporated within the same framework. Future work will also investigate approximation error and generalization properties across different metric families. These directions further improve the applicability of the framework for real-time robotic systems.

\bibliography{references.bib}

\begin{thebibliography}{10}
\providecommand{\url}[1]{#1}
\csname url@rmstyle\endcsname
\providecommand{\newblock}{\relax}
\providecommand{\bibinfo}[2]{#2}
\providecommand\BIBentrySTDinterwordspacing{\spaceskip=0pt\relax}
\providecommand\BIBentryALTinterwordstretchfactor{4}
\providecommand\BIBentryALTinterwordspacing{\spaceskip=\fontdimen2\font plus
\BIBentryALTinterwordstretchfactor\fontdimen3\font minus
  \fontdimen4\font\relax}
\providecommand\BIBforeignlanguage[2]{{%
\expandafter\ifx\csname l@#1\endcsname\relax
\typeout{** WARNING: IEEEtran.bst: No hyphenation pattern has been}%
\typeout{** loaded for the language `#1'. Using the pattern for}%
\typeout{** the default language instead.}%
\else
\language=\csname l@#1\endcsname
\fi
#2}}

\bibitem{411534}
A.~Bloch and P.~Croach, ``Reduction of euler lagrange problems for constrained
  variational problems and relation with optimal control problems,'' in
  \emph{Proceedings of 1994 33rd IEEE Conference on Decision and Control},
  vol.~3, 1994, pp. 2584--2590 vol.3.

\bibitem{crouch1995dynamic}
P.~Crouch and F.~S. Leite, ``The dynamic interpolation problem: on riemannian
  manifolds, lie groups, and symmetric spaces,'' \emph{Journal of Dynamical and
  control systems}, vol.~1, no.~2, pp. 177--202, 1995.

\bibitem{704225}
M.~Zefran, V.~Kumar, and C.~Croke, ``On the generation of smooth
  three-dimensional rigid body motions,'' \emph{IEEE Transactions on Robotics
  and Automation}, vol.~14, no.~4, pp. 576--589, 1998.

\bibitem{933127}
F.~Bullo and K.~Lynch, ``Kinematic controllability and decoupled trajectory
  planning for underactuated mechanical systems,'' in \emph{Proceedings 2001
  ICRA. IEEE International Conference on Robotics and Automation (Cat.
  No.01CH37164)}, vol.~4, 2001, pp. 3300--3307 vol.4.

\bibitem{1019463}
C.~Belta and V.~Kumar, ``An {SVD}-based projection method for interpolation on
  {$SE(3)$},'' \emph{IEEE Transactions on Robotics and Automation}, vol.~18,
  no.~3, pp. 334--345, 2002.

\bibitem{giambo2004optimal}
R.~Giambo, F.~Giannoni, and P.~Piccione, ``Optimal control on riemannian
  manifolds by interpolation,'' \emph{Mathematics of Control, Signals and
  Systems}, vol.~16, no.~4, pp. 278--296, 2004.

\bibitem{POPIEL2007111}
\BIBentryALTinterwordspacing
T.~Popiel and L.~Noakes, ``Bézier curves and c2 interpolation in riemannian
  manifolds,'' \emph{Journal of Approximation Theory}, vol. 148, no.~2, pp.
  111--127, 2007. [Online]. Available:
  \url{https://www.sciencedirect.com/science/article/pii/S0021904507000469}
\BIBentrySTDinterwordspacing

\bibitem{10.1093/imamci/12.4.399}
\BIBentryALTinterwordspacing
M.~Camarinha, F.~S. Leite, and P.~E. Crouch, ``Splines of class c k on
  non-euclidean spaces,'' \emph{IMA Journal of Mathematical Control and
  Information}, vol.~12, no.~4, pp. 399--410, 12 1995. [Online]. Available:
  \url{https://doi.org/10.1093/imamci/12.4.399}
\BIBentrySTDinterwordspacing

\bibitem{10.4310/CMS.2016.v14.n1.a3}
L.~Noakes and T.~Ratiu, ``Bi-jacobi fields and riemannian cubics for
  left-invariant {$SO(3)$},'' \emph{Communications in Mathematical Sciences},
  vol.~14, pp. 55--68, 01 2016.

\bibitem{DeCasteljau}
P.~Crouch, G.~Kun, and F.~Silva~Leite, ``The de casteljau algorithm on lie
  groups and spheres,'' \emph{Journal of Dynamical and Control Systems},
  vol.~5, pp. 397--429, 07 1999.

\bibitem{BoMoVe2018}
\BIBentryALTinterwordspacing
G.~Bogfjellmo, K.~Modin, and O.~Verdier, ``A numerical algorithm for
  {C2}-splines on symmetric spaces,'' \emph{SIAM J. Numer. Analysis}, vol.~56,
  no.~4, pp. 2623--2647, 2018. [Online]. Available:
  \url{https://doi.org/10.1137/17M1123353}
\BIBentrySTDinterwordspacing

\bibitem{watterson2018trajectory}
M.~Watterson, S.~Liu, K.~Sun, T.~Smith, and V.~Kumar, ``Trajectory optimization
  on manifolds with applications to {$SO(3)$} and {$R3 \times S2$} .'' in
  \emph{Robotics: Science and Systems}, vol.~20, 2018.

\bibitem{FLIESS01061995}
\BIBentryALTinterwordspacing
M.~Fliess, J.~Lévine, P.~Martin, and P.~Rouchon, ``Flatness and defect of
  non-linear systems: introductory theory and examples,'' \emph{International
  Journal of Control}, vol.~61, no.~6, pp. 1327--1361, 1995. [Online].
  Available: \url{https://doi.org/10.1080/00207179508921959}
\BIBentrySTDinterwordspacing

\bibitem{5980409}
D.~Mellinger and V.~Kumar, ``Minimum snap trajectory generation and control for
  quadrotors,'' in \emph{2011 IEEE International Conference on Robotics and
  Automation}, 2011, pp. 2520--2525.

\bibitem{7299672}
M.~W. Mueller, M.~Hehn, and R.~D'Andrea, ``A computationally efficient motion
  primitive for quadrocopter trajectory generation,'' \emph{IEEE Transactions
  on Robotics}, vol.~31, no.~6, pp. 1294--1310, 2015.

\bibitem{8206119}
S.~Liu, N.~Atanasov, K.~Mohta, and V.~Kumar, ``Search-based motion planning for
  quadrotors using linear quadratic minimum time control,'' in \emph{2017
  IEEE/RSJ International Conference on Intelligent Robots and Systems (IROS)},
  2017, pp. 2872--2879.

\bibitem{7128399}
M.~Hehn and R.~D’Andrea, ``Real-time trajectory generation for
  quadrocopters,'' \emph{IEEE Transactions on Robotics}, vol.~31, no.~4, pp.
  877--892, 2015.

\bibitem{10412114}
Y.~Wu, X.~Sun, I.~Spasojevic, and V.~Kumar, ``Deep learning for optimization of
  trajectories for quadrotors,'' \emph{IEEE Robotics and Automation Letters},
  vol.~9, no.~3, pp. 2479--2486, 2024.

\bibitem{9147300}
W.~Sun, G.~Tang, and K.~Hauser, ``Fast uav trajectory optimization using
  bilevel optimization with analytical gradients,'' in \emph{2020 American
  Control Conference (ACC)}, 2020, pp. 82--87.

\bibitem{7839930}
S.~Liu, M.~Watterson, K.~Mohta, K.~Sun, S.~Bhattacharya, C.~J. Taylor, and
  V.~Kumar, ``Planning dynamically feasible trajectories for quadrotors using
  safe flight corridors in 3-d complex environments,'' \emph{IEEE Robotics and
  Automation Letters}, vol.~2, no.~3, pp. 1688--1695, 2017.

\bibitem{tordesillas2021faster}
J.~Tordesillas and J.~P. How, ``{FASTER}: Fast and safe trajectory planner for
  navigation in unknown environments,'' \emph{IEEE Transactions on Robotics},
  2021.

\bibitem{5256286}
D.~Verscheure, B.~Demeulenaere, J.~Swevers, J.~De~Schutter, and M.~Diehl,
  ``Time-optimal path tracking for robots: A convex optimization approach,''
  \emph{IEEE Transactions on Automatic Control}, vol.~54, no.~10, pp.
  2318--2327, 2009.

\bibitem{5f032ddf-06b1-30a4-8397-578b4b7be0ad}
\BIBentryALTinterwordspacing
P.~E. Jupp and J.~T. Kent, ``Fitting smooth paths to speherical data,''
  \emph{Journal of the Royal Statistical Society. Series C (Applied
  Statistics)}, vol.~36, no.~1, pp. 34--46, 1987. [Online]. Available:
  \url{http://www.jstor.org/stable/2347843}
\BIBentrySTDinterwordspacing

\bibitem{4608619}
D.~Han, X.~Fang, and Q.~Wei, ``Rotation interpolation based on the geometric
  structure of unit quaternions,'' in \emph{2008 IEEE International Conference
  on Industrial Technology}, 2008, pp. 1--6.

\bibitem{BonalliBylardEtAl2019}
\BIBentryALTinterwordspacing
R.~Bonalli, A.~Bylard, A.~Cauligi, T.~Lew, and M.~Pavone, ``Trajectory
  optimization on manifolds: {A} theoretically-guaranteed embedded sequential
  convex programming approach,'' in \emph{{Robotics: Science and Systems}},
  Freiburg im Breisgau, Germany, June 2019. [Online]. Available:
  \url{https://arxiv.org/pdf/1905.07654.pdf}
\BIBentrySTDinterwordspacing

\bibitem{9765821}
Z.~Wang, X.~Zhou, C.~Xu, and F.~Gao, ``Geometrically constrained trajectory
  optimization for multicopters,'' \emph{IEEE Transactions on Robotics},
  vol.~38, no.~5, pp. 3259--3278, 2022.

\bibitem{10.1007/978-3-030-28619-4_20}
M.~Watterson and V.~Kumar, ``Control of quadrotors using the hopf fibration on
  {$SO(3)$},'' in \emph{Robotics Research}, N.~M. Amato, G.~Hager, S.~Thomas,
  and M.~Torres-Torriti, Eds.\hskip 1em plus 0.5em minus 0.4em\relax Cham:
  Springer International Publishing, 2020, pp. 199--215.

\bibitem{Watterson-IJRR-2020}
\BIBentryALTinterwordspacing
M.~Watterson, S.~Liu, K.~Sun, T.~Smith, and V.~Kumar, ``Trajectory optimization
  on manifolds with applications to quadrotor systems,'' \emph{The
  International Journal of Robotics Research}, vol.~39, no. 2-3, pp. 303--320,
  2020. [Online]. Available: \url{https://doi.org/10.1177/0278364919891775}
\BIBentrySTDinterwordspacing

\bibitem{teng2025riemannian}
S.~Teng, T.-Y. Lin, W.~A. Clark, R.~Vasudevan, and M.~Ghaffari, ``Riemannian
  direct trajectory optimization of rigid bodies on matrix lie groups,''
  \emph{arXiv preprint arXiv:2505.02323}, 2025.

\bibitem{song2023reaching}
Y.~Song, A.~Romero, M.~M{\"u}ller, V.~Koltun, and D.~Scaramuzza, ``Reaching the
  limit in autonomous racing: Optimal control versus reinforcement learning,''
  \emph{Science Robotics}, vol.~8, no.~82, p. eadg1462, 2023.

\bibitem{kaufmann2023champion}
E.~Kaufmann, L.~Bauersfeld, A.~Loquercio, M.~M{\"u}ller, V.~Koltun, and
  D.~Scaramuzza, ``Champion-level drone racing using deep reinforcement
  learning,'' \emph{Nature}, vol. 620, no. 7976, pp. 982--987, 2023.

\bibitem{loquercio2021learning}
A.~Loquercio, E.~Kaufmann, R.~Ranftl, M.~M{\"u}ller, V.~Koltun, and
  D.~Scaramuzza, ``Learning high-speed flight in the wild,'' \emph{Science
  Robotics}, vol.~6, no.~59, p. eabg5810, 2021.

\bibitem{8593536}
G.~Tang, W.~Sun, and K.~Hauser, ``Learning trajectories for real- time optimal
  control of quadrotors,'' in \emph{2018 IEEE/RSJ International Conference on
  Intelligent Robots and Systems (IROS)}, 2018, pp. 3620--3625.

\bibitem{pmlr-v205-ryou23a}
G.~Ryou, E.~Tal, and S.~Karaman, ``Real-time generation of time-optimal
  quadrotor trajectories with semi-supervised seq2seq learning,'' in
  \emph{Proceedings of The 6th Conference on Robot Learning}, ser. Proceedings
  of Machine Learning Research, K.~Liu, D.~Kulic, and J.~Ichnowski, Eds., vol.
  205.\hskip 1em plus 0.5em minus 0.4em\relax PMLR, 14--18 Dec 2023, pp.
  1860--1870.

\bibitem{pmlr-v211-tankasala23a}
\BIBentryALTinterwordspacing
S.~Tankasala and M.~Pryor, ``Accelerating trajectory generation for quadrotors
  using transformers,'' in \emph{Proceedings of The 5th Annual Learning for
  Dynamics and Control Conference}, ser. Proceedings of Machine Learning
  Research, N.~Matni, M.~Morari, and G.~J. Pappas, Eds., vol. 211.\hskip 1em
  plus 0.5em minus 0.4em\relax PMLR, 15--16 Jun 2023, pp. 600--611. [Online].
  Available: \url{https://proceedings.mlr.press/v211/tankasala23a.html}
\BIBentrySTDinterwordspacing

\bibitem{ryou2024multifidelityreinforcementlearningtimeoptimal}
\BIBentryALTinterwordspacing
G.~Ryou, G.~Wang, and S.~Karaman, ``Multi-fidelity reinforcement learning for
  time-optimal quadrotor re-planning,'' 2024. [Online]. Available:
  \url{https://arxiv.org/abs/2403.08152}
\BIBentrySTDinterwordspacing

\bibitem{mao2025sequence}
K.~Mao, H.~Yu, R.~Zhang, I.~Spasojevic, S.~Gao, and V.~Kumar, ``Sequence
  modeling for time-optimal quadrotor trajectory optimization with
  sampling-based robustness analysis,'' in \emph{Conference on Robot
  Learning}.\hskip 1em plus 0.5em minus 0.4em\relax PMLR, 2025, pp. 1531--1540.

\bibitem{han2025dynamically}
Z.~Han, L.~Xu, L.~Pei, and F.~Gao, ``Dynamically feasible trajectory generation
  with optimization-embedded networks for autonomous flight,'' \emph{IEEE
  Robotics and Automation Letters}, 2025.

\bibitem{10.5555/3305381.3305396}
B.~Amos and J.~Z. Kolter, ``Optnet: Differentiable optimization as a layer in
  neural networks,'' in \emph{Proceedings of the 34th International Conference
  on Machine Learning - Volume 70}, ser. ICML'17.\hskip 1em plus 0.5em minus
  0.4em\relax JMLR.org, 2017, p. 136–145.

\bibitem{agrawal2019differentiable}
A.~Agrawal, B.~Amos, S.~Barratt, S.~Boyd, S.~Diamond, and J.~Z. Kolter,
  ``Differentiable convex optimization layers,'' \emph{Advances in neural
  information processing systems}, vol.~32, 2019.

\bibitem{donti_dc3_2021}
P.~Donti, D.~Rolnick, and J.~Z. Kolter, ``Dc3: A learning method for
  optimization with hard constraints,'' in \emph{International Conference on
  Learning Representations}, 2021.

\bibitem{negiar_learning_2023}
G.~N{\'e}giar, M.~W. Mahoney, and A.~Krishnapriyan, ``Learning differentiable
  solvers for systems with hard constraints,'' in \emph{The Eleventh
  International Conference on Learning Representations}, 2023.

\bibitem{min2024hardconstrainedneuralnetworksuniversal}
\BIBentryALTinterwordspacing
Y.~Min, A.~Sonar, and N.~Azizan, ``Hard-constrained neural networks with
  universal approximation guarantees,'' 2024. [Online]. Available:
  \url{https://arxiv.org/abs/2410.10807}
\BIBentrySTDinterwordspacing

\bibitem{RAISSI2019686}
\BIBentryALTinterwordspacing
M.~Raissi, P.~Perdikaris, and G.~Karniadakis, ``Physics-informed neural
  networks: A deep learning framework for solving forward and inverse problems
  involving nonlinear partial differential equations,'' \emph{Journal of
  Computational Physics}, vol. 378, pp. 686--707, 2019. [Online]. Available:
  \url{https://www.sciencedirect.com/science/article/pii/S0021999118307125}
\BIBentrySTDinterwordspacing

\bibitem{doi:10.1137/21M1397908}
\BIBentryALTinterwordspacing
L.~Lu, R.~Pestourie, W.~Yao, Z.~Wang, F.~Verdugo, and S.~G. Johnson,
  ``Physics-informed neural networks with hard constraints for inverse
  design,'' \emph{SIAM Journal on Scientific Computing}, vol.~43, no.~6, pp.
  B1105--B1132, 2021. [Online]. Available:
  \url{https://doi.org/10.1137/21M1397908}
\BIBentrySTDinterwordspacing

\bibitem{SON2023126424}
\BIBentryALTinterwordspacing
H.~Son, S.~W. Cho, and H.~J. Hwang, ``Enhanced physics-informed neural networks
  with augmented lagrangian relaxation method (al-pinns),''
  \emph{Neurocomputing}, vol. 548, p. 126424, 2023. [Online]. Available:
  \url{https://www.sciencedirect.com/science/article/pii/S0925231223005477}
\BIBentrySTDinterwordspacing

\bibitem{jaquier_learning_2022}
N.~Jaquier, Y.~Zhou, J.~Starke, and T.~Asfour, ``Learning to sequence and blend
  robot skills via differentiable optimization,'' \emph{IEEE Robotics and
  Automation Letters}, vol.~7, no.~3, pp. 8431--8438, 2022.

\bibitem{amos2018differentiable}
B.~Amos, I.~D.~J. Rodriguez, J.~Sacks, B.~Boots, and J.~Z. Kolter,
  ``Differentiable mpc for end-to-end planning and control,'' in
  \emph{Proceedings of the 32nd International Conference on Neural Information
  Processing Systems}, ser. NIPS'18.\hskip 1em plus 0.5em minus 0.4em\relax Red
  Hook, NY, USA: Curran Associates Inc., 2018, p. 8299–8310.

\bibitem{tordesillas2023rayenimpositionhardconvex}
\BIBentryALTinterwordspacing
J.~Tordesillas, J.~P. How, and M.~Hutter, ``Rayen: Imposition of hard convex
  constraints on neural networks,'' 2023. [Online]. Available:
  \url{https://arxiv.org/abs/2307.08336}
\BIBentrySTDinterwordspacing

\bibitem{duong21hamiltonian}
T.~Duong and N.~Atanasov, ``{Hamiltonian-based Neural ODE Networks on the SE(3)
  Manifold For Dynamics Learning and Control},'' in \emph{Proceedings of
  Robotics: Science and Systems}, Virtual, July 2021.

\bibitem{hutchinson2021lietransformer}
M.~J. Hutchinson, C.~Le~Lan, S.~Zaidi, E.~Dupont, Y.~W. Teh, and H.~Kim,
  ``Lietransformer: equivariant self-attention for lie groups,'' in
  \emph{International Conference on Machine Learning}.\hskip 1em plus 0.5em
  minus 0.4em\relax PMLR, 2021, pp. 4533--4543.

\bibitem{esteves2018learning}
C.~Esteves, C.~Allen-Blanchette, A.~Makadia, and K.~Daniilidis, ``Learning so
  (3) equivariant representations with spherical cnns,'' in \emph{Proceedings
  of the European Conference on Computer Vision (ECCV)}, 2018, pp. 52--68.

\bibitem{Arvanitidis2021_1000137507}
H.~Beik-Mohammadi, S.~Hauberg, G.~Arvanitidis, G.~Neumann, and L.~Rozo,
  ``\BIBforeignlanguage{english}{Learning riemannian manifolds for geodesic
  motion skills},'' in \emph{\BIBforeignlanguage{english}{Robotics: Science and
  System XVII, July 12 - July 16, 2021 --Held Virtually--, Ed.: D. A. Shell}},
  2021.

\bibitem{arvanitidis2019fast}
G.~Arvanitidis, S.~Hauberg, P.~Hennig, and M.~Schober, ``Fast and robust
  shortest paths on manifolds learned from data,'' in \emph{The 22nd
  International Conference on Artificial Intelligence and Statistics}.\hskip
  1em plus 0.5em minus 0.4em\relax PMLR, 2019, pp. 1506--1515.

\bibitem{11128556}
M.~D. Vedove, F.~J. Abu-Dakka, L.~Palopoli, D.~Fontanelli, and M.~Saveriano,
  ``Meshdmp: Motion planning on discrete manifolds using dynamic movement
  primitives,'' in \emph{2025 IEEE International Conference on Robotics and
  Automation (ICRA)}, 2025, pp. 895--901.

\bibitem{pmlr-v229-lee23a}
\BIBentryALTinterwordspacing
B.~Lee, Y.~Lee, S.~Kim, M.~Son, and F.~C. Park, ``Equivariant motion manifold
  primitives,'' in \emph{Proceedings of The 7th Conference on Robot Learning},
  ser. Proceedings of Machine Learning Research, J.~Tan, M.~Toussaint, and
  K.~Darvish, Eds., vol. 229.\hskip 1em plus 0.5em minus 0.4em\relax PMLR,
  06--09 Nov 2023, pp. 1199--1221. [Online]. Available:
  \url{https://proceedings.mlr.press/v229/lee23a.html}
\BIBentrySTDinterwordspacing

\bibitem{10637485}
Y.~Lee, ``Mmp++: Motion manifold primitives with parametric curve models,''
  \emph{IEEE Transactions on Robotics}, vol.~40, pp. 3950--3963, 2024.

\bibitem{10292912}
P.~Kicki, P.~Liu, D.~Tateo, H.~Bou-Ammar, K.~Walas, P.~Skrzypczyński, and
  J.~Peters, ``Fast kinodynamic planning on the constraint manifold with deep
  neural networks,'' \emph{IEEE Transactions on Robotics}, vol.~40, pp.
  277--297, 2024.

\bibitem{do1992riemannian}
M.~P. Do~Carmo and J.~Flaherty~Francis, \emph{Riemannian geometry}.\hskip 1em
  plus 0.5em minus 0.4em\relax Springer, 1992, vol.~2.

\bibitem{doi:10.1177/027836499901800208}
\BIBentryALTinterwordspacing
M.~Žefran, V.~Kumar, and C.~Croke, ``Metrics and connections for rigid-body
  kinematics,'' \emph{The International Journal of Robotics Research}, vol.~18,
  no.~2, pp. 242--1--242--16, 1999. [Online]. Available:
  \url{https://doi.org/10.1177/027836499901800208}
\BIBentrySTDinterwordspacing

\bibitem{doi:10.1177/027836499401300101}
\BIBentryALTinterwordspacing
F.~C. Park and R.~W. Brockett, ``Kinematic dexterity of robotic mechanisms,''
  \emph{The International Journal of Robotics Research}, vol.~13, no.~1, pp.
  1--15, 1994. [Online]. Available:
  \url{https://doi.org/10.1177/027836499401300101}
\BIBentrySTDinterwordspacing

\bibitem{geist2023rotations}
A.~R. Geist, J.~Frey, M.~Zhobro, A.~Levina, and G.~Martius, ``Learning with
  3{D} rotations, a hitchhiker’s guide to {SO}(3),'' in \emph{Proceedings of
  the 41st International Conference on Machine Learning}, ser. Proceedings of
  Machine Learning Research, vol. 235.\hskip 1em plus 0.5em minus 0.4em\relax
  PMLR, 21--27 Jul 2024, pp. 15\,331--15\,350.

\bibitem{doi:10.1137/17M1129416}
\BIBentryALTinterwordspacing
E.~S. Gawlik and M.~Leok, ``Embedding-based interpolation on the special
  orthogonal group,'' \emph{SIAM Journal on Scientific Computing}, vol.~40,
  no.~2, pp. A721--A746, 2018. [Online]. Available:
  \url{https://doi.org/10.1137/17M1129416}
\BIBentrySTDinterwordspacing

\bibitem{richter2016polynomial}
C.~Richter, A.~Bry, and N.~Roy, ``Polynomial trajectory planning for aggressive
  quadrotor flight in dense indoor environments,'' in \emph{Robotics Research:
  The 16th International Symposium ISRR}.\hskip 1em plus 0.5em minus
  0.4em\relax Springer, 2016, pp. 649--666.

\bibitem{2020SciPy-NMeth}
\BIBentryALTinterwordspacing
P.~Virtanen, R.~Gommers, T.~E. Oliphant, M.~Haberland, T.~Reddy, D.~Cournapeau,
  E.~Burovski, P.~Peterson, W.~Weckesser, J.~Bright, S.~J. van~der Walt,
  M.~Brett, J.~Wilson, K.~J. Millman, N.~Mayorov, A.~R.~J. Nelson, E.~Jones,
  R.~Kern, E.~Larson, C.~Carey, İbrahim Polat, Y.~Feng, E.~W. Moore,
  J.~VanderPlas, D.~Laxalde, J.~Perktold, R.~Cimrman, I.~Henriksen, E.~A.
  Quintero, C.~R. Harris, A.~M. Archibald, A.~H. Ribeiro, F.~Pedregosa, P.~van
  Mulbregt, and S.~. Contributors, ``{SciPy} 1.0: Fundamental algorithms for
  scientific computing in python,'' \emph{Nature Methods}, vol.~17, pp.
  261--272, 2020. [Online]. Available:
  \url{https://www.nature.com/articles/s41592-019-0686-2}
\BIBentrySTDinterwordspacing

\bibitem{9561840}
L.~Jarin-Lipschitz, J.~Paulos, R.~Bjorkman, and V.~Kumar,
  ``Dispersion-minimizing motion primitives for search-based motion planning,''
  in \emph{2021 IEEE International Conference on Robotics and Automation
  (ICRA)}, 2021, pp. 12\,625--12\,631.

\bibitem{11015263}
J.~Hou, X.~Zhou, N.~Pan, A.~Li, Y.~Guan, C.~Xu, Z.~Gan, and F.~Gao,
  ``Primitive-swarm: An ultra-lightweight and scalable planner for large-scale
  aerial swarms,'' \emph{IEEE Transactions on Robotics}, vol.~41, pp.
  3629--3648, 2025.

\bibitem{folk2023rotorpy}
S.~Folk, J.~Paulos, and V.~Kumar, ``{RotorPy}: A python-based multirotor
  simulator with aerodynamics for education and research,'' \emph{arXiv
  preprint arXiv:2306.04485}, 2023.

\bibitem{11474851}
K.~Kondo, Y.~Wu, V.~Kumar, and J.~P. How, ``Mighty: Hermite spline-based
  efficient trajectory planning,'' \emph{IEEE Robotics and Automation Letters},
  vol.~11, no.~6, pp. 6664--6671, 2026.

\end{thebibliography}

\end{document}